\documentclass[sigconf]{acmart}
\AtBeginDocument{%
  }

\setcopyright{acmlicensed}
\copyrightyear{2018}
\acmYear{2018}
\acmDOI{XXXXXXX.XXXXXXX}
\acmISBN{978-1-4503-XXXX-X/2018/06}

\usepackage{bm}
\usepackage{algorithm,algorithmic}
\usepackage{multirow}
\usepackage{subcaption}
\usepackage{diagbox}
\usepackage{pifont}
\usepackage{makecell}
\usepackage{arydshln}
\usepackage{graphicx}
\usepackage{booktabs}
\usepackage{subcaption}
\usepackage{enumitem}
\usepackage{xr-hyper}  % 外部引用支持 hyperref
\usepackage{subfiles}
\begin{document}

\title{Small Shifts, Large Gains: Unlocking Traditional TSP Heuristic Guided-Sampling via Unsupervised Neural Instance Modification}


\author{Wei Huang}
\affiliation{%
  \institution{University of New South Wales}
  \country{Australia}
}
\email{w.c.huang@unsw.edu.au}

\author{Hanchen Wang}
\affiliation{%
  \institution{University of Technology Sydney}
  \country{Australia}
}
\email{Hanchen.Wang@uts.edu.au}

\author{Dong Wen}
\affiliation{%
  \institution{University of New South Wales}
  \country{Australia}
}
\email{dong.wen@unsw.edu.au}

\author{Wenjie Zhang}
\affiliation{%
  \institution{University of New South Wales}
  \country{Australia}
}
\email{wenjie.zhang@unsw.edu.au}

\begin{abstract}
  The Traveling Salesman Problem (TSP) is one of the most representative NP-hard problems in route planning and a long-standing benchmark in combinatorial optimization. Traditional heuristic tour constructors, such as Farthest or Nearest Insertion, are computationally efficient and highly practical, but their deterministic behavior limits exploration and often leads to local optima. In contrast, neural-based heuristic tour constructors alleviate this issue through guided-sampling and typically achieve superior solution quality, but at the cost of extensive training and reliance on ground-truth supervision, hindering their practical use. To bridge this gap, we propose TSP-MDF, a novel instance modification framework that equips traditional deterministic heuristic tour constructors with guided-sampling capability. Specifically, TSP-MDF introduces a neural-based instance modifier that strategically shifts node coordinates to sample multiple modified instances, on which the base traditional heuristic tour constructor constructs tours that are mapped back to the original instance, allowing traditional tour constructors to explore higher-quality tours and escape local optima. At the same time, benefiting from our instance modification formulation, the neural-based instance modifier can be trained efficiently without any ground-truth supervision, ensuring the framework maintains practicality. Extensive experiments on large-scale TSP benchmarks and real-world benchmarks demonstrate that TSP-MDF significantly improves the performance of traditional heuristics tour constructors, achieving solution quality comparable to neural-based heuristic tour constructors, but with an extremely short training time. 
\end{abstract}





\maketitle
\begin{figure}[t]
     \centering
     \resizebox{0.15\textwidth}{!}{\includegraphics[scale=1]{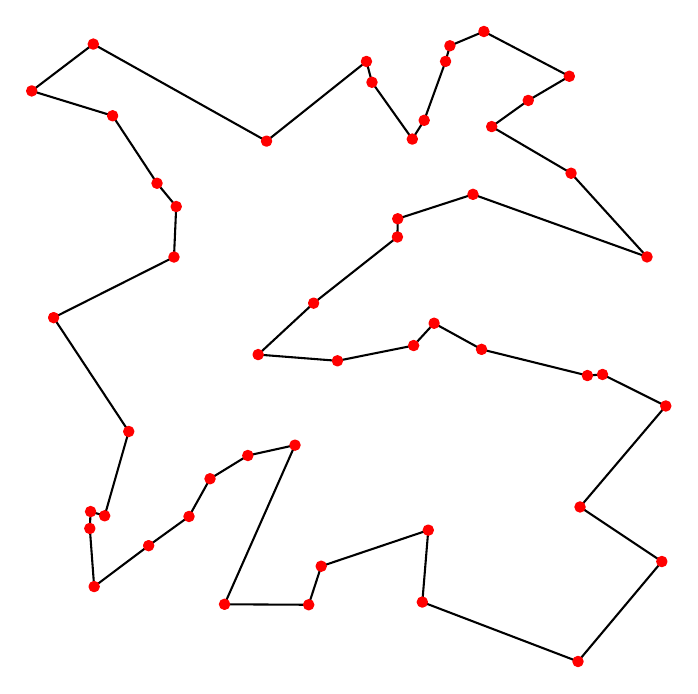}}
     \caption{An example of an optimal tour for a TSP instance.}
     \label{fig:example}
     \vspace{-0.5cm}
\end{figure}
\section{Introduction}
The Traveling Salesman Problem (TSP), one of the most representative NP-hard problems, aims to find a shortest cycle that visits each node exactly once, given a set of nodes in Euclidean space as shown in Figure~\ref{fig:example}. Its fundamental nature and wide-ranging applications, from route planning to circuit design \cite{vesselinova2020learning,li2023finding,hacizade2018ga}, make it a central problem in combinatorial optimization.%, and extensive efforts have been made to developing efficient and effective approximation methods. 
Existing approximation methods for TSP usually adopt a two-stage strategy: (1) a tour construction stage, where traditional or neural-based heuristic tour constructors are employed to construct one or multiple candidate tours \cite{qiu2022dimes,sun2023difusco,li2024fast,kwon2020pomo,joshi2020learning,jin2023pointerformer,li2023t2t,hottung2021efficient,fu2021generalize,kim2021learning,kool2018attention,ye2023deepaco,li2025diversity,goh2024hierarchical}; (2) an optional tour refinement stage that iteratively improves the constructed candidate tours refined~\cite{croes1958method,ye2024glop,kim2021learning,fu2021generalize,helsgaun2017extension,ye2023deepaco}.

Regarding the tour construction stage, which has been the main focus of existing works, traditional heuristic tour constructors often construct a tour by iteratively inserting nodes into the partial tour, where the insertion order is determined by hand-crafted rules (e.g., Farthest/Nearest Insertion) based on the pairwise distances in the underlying TSP instance. 
These heuristics are efficient (e.g., $O(n^2)$) and often yield an initial tour of reasonable quality. However, they often struggle in certain complex regions (e.g., dense clusters or distant outliers) of the underlying TSP instance, where finding a sub-tour that aligns with the global optimum is inherently difficult, and a sub-optimal sub-tour can propagate their effects to the global level, \textbf{leading to a sub-optimal overall tour}.
Although the subsequent tour refinement stage can refine the constructed tour via local search~\cite{croes1958method,ye2024glop,kim2021learning,helsgaun2017extension,ye2023deepaco}, \textbf{unfortunately, its effectiveness heavily depends on the quality of the initial tour.} When the initial tour is of poor quality, local search may converge to another undesirable sub-optimal solution within the same local optimum, and may also require a large number of refinement steps, resulting in high computational cost compared to initial tour construction. Thus, \textbf{constructing high quality initial tours remains crucial for effective TSP approximation}.

A common approach to escape local optima is to sample multiple candidate initial tours during tour construction stage \cite{ye2024glop,kim2021learning,helsgaun2017extension}, enabling broader exploration of the solution space, especially for large instances with extremely large solution space. However, most traditional heuristic constructors (e.g., Farthest/Nearest Insertion) \textbf{are inherently deterministic and typically produce only a single tour for a given instance}.
Therefore, multiple candidates are often obtained via random insertion orders or random tours in practice, which leads to unstable tour quality and requires extensive sampling, resulting in inefficient exploration.
And designing effective heuristics to overcome these limitations usually demands substantial manual effort.

To address these limitations, neural-based heuristic tour constructors have been widely studied~\cite{qiu2022dimes,sun2023difusco,li2024fast,li2023t2t,hottung2021efficient,fu2021generalize,kwon2020pomo,joshi2020learning,kim2021learning,jin2023pointerformer,kool2018attention, ye2023deepaco,li2025diversity,goh2024hierarchical}. They have achieved promising results by performing guided-sampling using heuristics automatically learned by neural networks, and are often more flexible than traditional counterparts. Specifically, given a TSP instance, they estimate distributions over promising regions of the solution space, enabling the sampling of multiple candidate tours with controlled quality. Such guided-sampling improves exploration, increases the likelihood of escaping local optima and discovering higher-quality tours.

However, existing neural-based heuristic tour constructors suffer from several practical issues: (1) they often \textbf{require large amounts of ground-truth training data}, which is costly to obtain in practice, especially for large instances; (2) Although there exist unsupervised reinforcement learning approaches \cite{kwon2020pomo,qiu2022dimes,jin2023pointerformer,hottung2021efficient,joshi2020learning,kim2021learning,kool2018attention, ye2023deepaco,li2025diversity,goh2024hierarchical} that avoid the need for labeled data, they generally \textbf{produce inferior solutions} due to the large complex solution space and the low sampling efficiency of reinforcement learning; (3) As neural-based heuristic tour constructors have to be trained entirely from scratch, they often learn ineffective and unstable heuristics that \textbf{yield poor solutions in the early training stage}. 
And since the high-quality solution space is extremely small and highly constrained, occupying only a tiny fraction of the overall solution space, learning effective heuristics becomes particularly challenging and often \textbf{requires an extremely long training time} to achieve satisfactory performance, even on small instances.
These limitations collectively increase training cost, thereby \textbf{restricting the practicability of neural-based heuristic tour constructors in real-world scenarios}, particularly under limited computational budgets. In contrast, traditional heuristic constructors can directly produce reasonable initial tours without training or labeled data. This motivates a key question for tour construction: \textbf{\textit{``Instead of learning a neural heuristic from scratch, can we leverage a traditional heuristic tour constructor as a practical base constructor, and enhance it by equipping it with guided-sampling capability, while preserving practicability?''}}

\begin{figure*}[ht]
     \centering
     \resizebox{\textwidth}{!}{\includegraphics[scale=1]{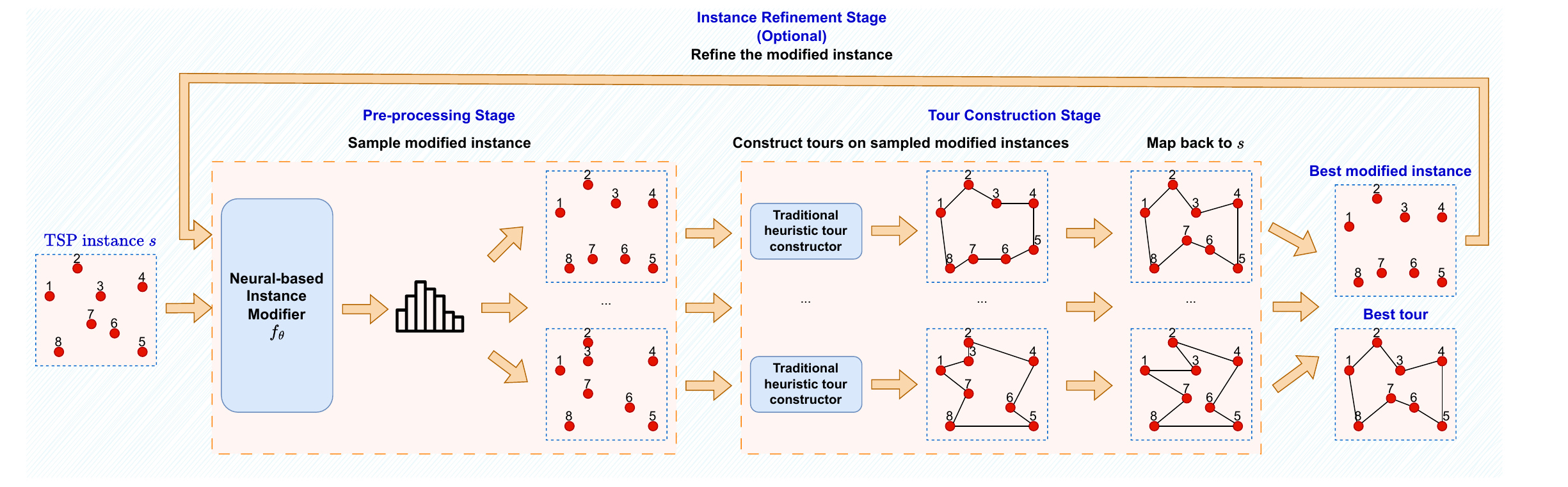}}
     \caption{An overview of our TSP-MDF. In the pre-processing stage, we sample multiple modified instances by leveraging a neural-based instance modifier to modify the coordinates of nodes in the original instance. In the tour construction stage, traditional heuristic tour constructors are executed on these modified instances to construct candidate tours, which are then mapped back to the original instance. In the optional instance refinement stage, the modified instance that currently results in the best tour can be further modified to sample new tours.}
     \label{fig:tsp_mdf}
     \vspace{-0.3cm}
\end{figure*}

To answer this question, we propose TSP-MDF, a new formulation for TSP tour construction. Unlike prior work that redesigns heuristics, TSP-MDF introduces an instance modification framework that enables traditional deterministic heuristic constructors to perform guided sampling by approximating tours on multiple modified instances.
In particular, we introduce an additional pre-processing stage before the tour construction stage as shown in Figure \ref{fig:tsp_mdf}, where an unsupervised neural-based instance modifier is trained to strategically shift the node coordinates of the original TSP instance and sample multiple modified instances. Then, by simply applying the traditional heuristic tour constructor to these modified instances, we can obtain multiple candidate tours and map them back to the original instance. 
In this way, \textbf{traditional deterministic heuristic tour constructors are now equipped with guided-sampling capability}. 
Moreover, benefiting from our formulation, TSP-MDF also introduces an optional instance refinement stage, where modified instances can be iteratively adjusted to further guide the heuristic constructors toward better tours.
To this end, TSP-MDF provides great flexibility for traditional deterministic heuristic tour constructors to escape local optima and explore higher-quality tours by \textbf{leveraging both parallel and sequential guided-sampling}.

Crucially, the space of high-quality instance modifications is far less constrained than that of high-quality tours, as multiple effective modifications may lead to the same improved tour.  
Meanwhile, this formulation allows a large portion of the unpromising solution space to be quickly pruned by restricting modifications to remain close to the original instance, substantially increasing the density of high-quality modifications in the search space. This significantly reduces the difficulty of finding effective modifications, thus making our neural-based instance modifier practical, it can \textbf{achieve satisfactory performance with remarkably short training time and without requiring any ground-truth training data}.

Our main contributions can be summarized as follows:
\begin{itemize}[leftmargin=*]
    
    \item We propose TSP-MDF, providing a novel perspective for TSP tour construction that leverages a traditional deterministic heuristic tour constructor as base tour constructor and introduces a pre-processing stage to sample multiple modified instances. By approximating tours on these sampled instances, TSP-MDF unlocks the guided-sampling capability of the base tour constructor, thereby enhancing solution quality.

    \item Our formulation further introduces an optional instance refinement stage, which iteratively adjusts the modified instances to guide traditional heuristic tour constructors toward better tours, providing greater flexibility in escaping local optima and discovering higher-quality solutions through both parallel and sequential guided-sampling.
        
    \item We design a practical unsupervised neural-based TSP instance modifier that can effectively guide the sampling of modified instances, while eliminating the need for ground-truth supervision and costly training.

    \item Extensive experiments on large-scale and real-world benchmarks show that TSP-MDF significantly improves the solution quality of traditional deterministic heuristic tour constructors, achieving consistent performance comparable to neural-based heuristic tour constructors, but with extremely short training time and without requiring any ground-truth supervision. These demonstrate that our framework is both effective, efficient, and practical.
\end{itemize}

\section{Related Works}
\noindent\textbf{Exact methods for TSP.}\quad
Traditional exact solvers for the TSP~\cite{gurobi,applegate2006concorde,cplex2009v12} are mainly built upon linear programming and mixed integer linear programming frameworks. Among them, Concorde~\cite{applegate2006concorde} is one of the most famous state-of-the-art exact solvers
based on an advanced branch-and-cut algorithm. Despite their strong optimality guarantees, these exact solvers are computationally expensive and become impractical for large-scale TSP instances.

\noindent\textbf{Approximation methods for TSP.}\quad
To improve efficiency, TSP solutions are commonly approximated using a two-stage framework: an initial tour construction stage that generates one or more candidate tours, followed by an optional tour refinement stage to improve solution quality.

In the construction stage, classical heuristics such as Farthest and Nearest Insertion iteratively build tours by inserting nodes according to distance-based rules. These methods are computationally efficient and produce reasonable solutions, but are inherently deterministic and typically generate only a single tour per instance, limiting exploration.
Recently, neural-based heuristic tour constructors~\cite{qiu2022dimes,sun2023difusco,li2024fast,kwon2020pomo,joshi2020learning,jin2023pointerformer,li2023t2t,hottung2021efficient,fu2021generalize,kim2021learning,kool2018attention, ye2023deepaco,li2025diversity,goh2024hierarchical} have emerged as popular alternatives for constructing TSP tours. These methods aim to estimate distributions that highlight promising regions of the solution space, from which multiple candidate tours can be sampled and explored.
Early neural-based frameworks~\cite{qiu2022dimes,kwon2020pomo,joshi2020learning,jin2023pointerformer,hottung2021efficient,kim2021learning,kool2018attention} primarily focus on developing unsupervised reinforcement learning methods to sequentially sample and insert nodes into partial tours. While effective on small-scale instances, these approaches often struggle to scale to larger instances due to the combinatorial complexity of the solution space and the low sampling efficiency of reinforcement learning.
More recently, generative diffusion-based models~\cite{sun2023difusco,li2023t2t,li2024fast} trained with supervised learning have demonstrated promising performance on large-scale TSP instances. By generating multiple diverse and high-quality distributions for a given instance, these models are able to effectively capture the multimodal nature of the problem. However, they rely heavily on supervised training and require access to high-quality ground-truth solutions, which are computationally expensive to obtain in practice. Moreover, existing neural-based heuristic tour constructors typically need to be trained entirely from scratch and often incur substantial training costs before achieving satisfactory performance, even on relatively small instances, which limits their practicality in real-world settings.

Notably, in this paper, we focus on the tour construction stage, due to the space limitation, an extended related work for the tour refinement stage can be found in Appendix~\ref{appendix:related_works}.

\section{Preliminary}
In this paper, we focus on the classic 2-D Euclidean TSP, which has been the
most intensely studied setting in most previous works~\cite{qiu2022dimes,sun2023difusco,li2024fast,kwon2020pomo,joshi2020learning,jin2023pointerformer,li2023t2t,hottung2021efficient,fu2021generalize,kim2021learning,kool2018attention,ye2023deepaco,li2025diversity,goh2024hierarchical,ye2024glop}.

\noindent\textbf{Traveling Salesman Problem (TSP).} \textit{Given a TSP instance $s=\{v^s_1, v^s_2, \dots, v^s_n\}$ with a set of $n$ nodes in the $2$-dimensional Euclidean space, where each node $i$ has coordinates $v^s_i = (v^s_{i,1}, v^s_{i,2}) \in \mathbb{R}^2$, find a tour $\pi$ (i.e., a permutation of $\{1, 2, \dots, n\}$) that minimizes the total tour length on $s$:
\begin{equation}
    c_s({\pi}) = \sum_{i=1}^{n}\|v^s_{\pi_i}-v^s_{\pi_{i+1}}\|_2 , \quad \text{with } \pi_{n+1} = \pi_1
\end{equation}
where $\|v^s_{\pi_i}-v^s_{\pi_{i+1}}\|_2$ is the Euclidean distance between $v^s_{\pi_i}$ and $v^s_{\pi_{i+1}}$.}

\section{Proposed Approach: TSP-MDF}
\subsection{TSP-MDF Overview}
In this section, we present an overview of TSP-MDF, which leverages an existing practical traditional deterministic heuristic tour constructor as the base tour constructor, and aims to 
address the question: \textbf{\textit{``Can we enhance the base tour constructor by unlocking its guided-sampling capability to approximate multiple alternative better tours, while ensuring practicability?"}}

To address the above question, we first tackle a key sub-question: \textbf{\textit{``How can we guide a traditional deterministic heuristic tour constructor to approximate alternative tours that have the potential to outperform its original tour?"}}.
Inspired by prompt engineering in Large Language Models (LLMs), where input prompts are carefully modified to guide the model toward producing better responses, our TSP-MDF introduces an additional pre-processing stage before the initial tour construction stage. Specifically, given an original TSP instance $s$, our pre-processing stage strategically modifies $s$ by shifting node coordinates to sample a set of modified instances $S_{1}$. Each modified instance $s_1 \in S_1$ aims to guide the traditional heuristic tour constructor in the subsequent tour construction stage to approximate a tour $\pi^{s_1}=\mathrm{Heuristic}(s_1)$ on $s_1$, which can then be mapped back to the original instance $s$ to yield an alternative tour of potentially shorter length $c_s(\pi^{s_1})$ on $s$, as illustrated in Figure~\ref{fig:tsp_mdf}. This approach does not require re-designing the traditional heuristic tour constructors, and is reasonable because the primary challenge of the TSP often lies in certain complex regions of the instance, where identifying a sub-tour that aligns with the global optimum is inherently difficult for traditional heuristic tour constructors. By strategically modifying node coordinates to adjust the relative distances among nodes, these complex regions can be simplified, enabling traditional heuristic tour constructors to more easily identify a better sub-tour that is otherwise difficult to obtain in the original instance, ultimately leading to an improved global tour on the original instance.

Moreover, directly obtaining high-quality modifications can be challenging. Fortunately, our pre-processing stage is not limited to modifying only the original instance, it can be also applied to further modify previously modified instances. This design offers an extra optional way to guide the traditional heuristic tour constructor towards approximating better tours by iteratively refining an initial instance $s$ for an arbitrary number of iterations. Specifically, at each iteration $t$, given an instance $s_t$ obtained after $t$ modifications, we can further refine $s_t$ using the pre-processing stage to sample a set of new instances $S_{t+1}$, from which multiple new tours can be obtained. Thus, the overall pipeline of TSP-MDF can be viewed as an auto-regressive process starts from $s_0=s$. Note that, in practice, to sample a set of modified instances $S_{t+1}$ at each iteration $t$, instead of modifying every $s_t \in S_t$, we adopt a greedy strategy for simplicity and efficiency. Specifically, we sample $S_{t+1}$ by only modifying the current best modified instance $s^*_t$, which yields the shortest tour length on $s$ so far, such that:
\begin{equation}
\label{eq:s_best}
\begin{split}
    s^*_t &= \arg\min_{s'\in\{s^*_{t-1}\}\cup S_t} c_s(\pi^{s'}),\quad\mathrm{with}\quad s^*_0 =s
\end{split}
\end{equation}

To this end, TSP-MDF enables both parallel and sequential guided-sampling for traditional heuristic constructors, and this brings to another key sub-question: \textbf{\textit{``How can we strategically modify an instance and sample multiple modified instances, while ensuring practicability?''}}.
Similar to the neural-based heuristic tour constructors, we introduce a neural-based instance modifier $f_\theta$ that automatically learns to modify the instance and estimates a distribution over modified instances. Sampling from this distribution produces multiple candidates, each capable of guiding the traditional heuristic constructor toward improved tours.
Notably, the traditional heuristic tour constructor can be applied to all modified instances in parallel at each iteration, thus the additional computational overhead of TSP-MDF primarily arises from the sampling of modified instances during the pre-processing stage. Since these instances are sampled simultaneously from a distribution generated by a single forward pass of the neural instance modifier, the overall inference time of TSP-MDF remains comparable to that of neural-based heuristic tour constructors.

Importantly, our neural-based instance modifier offers a potential practical advantage over neural-based heuristic tour constructors in terms of learning difficulty. This is because the space of high-quality tours is extremely small and highly constrained, only a few high-quality tours may exist for a given instance, making it challenging for neural-based heuristic tour constructors to precisely discover such tours, thus requiring substantial training time. In contrast, under our formulation, a high-quality tour can correspond to multiple distinct instance modifications, so the neural-based instance modifier only needs to discover any one of these effective modifications to guide the base tour constructor toward an improved tour, which substantially increases the tolerance to modeling errors and the probability of success, thereby potentially reducing the overall learning difficulty.
Finally, our goal is to design an unsupervised neural-based instance modifier that can effectively modify instances, but without requiring costly training.

\subsection{Basic Design of $f_\theta$}
\label{sec:basic}
In this section, we present a basic design of the neural-based instance modifier $f_\theta$, including its sampling strategy and unsupervised training strategy.

\vspace{1mm}\noindent\textbf{Modified instance sampling.}\quad 
In each modification iteration of TSP-MDF, given an original TSP instance $s$ and a current best modified instance $s^*_{t}$, the neural-based instance modifier aims to modify $s^*_{t}$ to sample a set of modified instances $S_{t+1}$, where each $s_{t+1} \in S_{t+1}$ contains a set of $n$ nodes $s_{t+1}=\{v_1^{s_{t+1}},\dots,v_n^{s_{t+1}}\}$, and each node $i$ consists of two modified coordinates $v_i^{s_{t+1}}=(v_{i,1}^{s_{t+1}},v_{i,2}^{s_{t+1}})$. A basic design is then to take both $s$ and $s^*_{t}$ as inputs, and learn to modify the instance by estimating a distribution $f^{\Delta v_{i,k}^{s_{t+1}}}_\theta(\cdot|s^*_{t},s)$ for each coordinate dimension $k$ of each node $i$, from which a refined coordinate offset $\Delta v_{i,k}^{s_{t+1}}$ can be sampled. 

A modified instance $s_{t+1}$ can then be sampled by sampling each new coordinate offset $\Delta v_{i,k}^{s_{t+1}}$ from its corresponding distribution and adding it to the corresponding original coordinate in $s$:
\begin{equation}
\label{eq:delta_sampling}
\begin{split}
    \Delta v_{i,k}^{s_{t+1}} &\sim f^{\Delta v_{i,k}^{s_{t+1}}}_\theta(\cdot | s^*_{t}, s), 
    \quad i=1,\dots,n,\; k=1,2 \\
    s_{t+1} = s + \Delta s_{t+1},\quad& \Delta s_{t+1} = \{(\Delta v_{1,1}^{s_{t+1}},\Delta v_{1,2}^{s_{t+1}}),\dots,(\Delta v_{n,1}^{s_{t+1}},\Delta v_{n,2}^{s_{t+1}})\}\\
\end{split}
\end{equation}
where $\Delta s_{t+1}$ denotes the overall instance modification, and its overall probability (density) can be expressed as:
\begin{equation}
\label{eq:prob_basic}
\begin{split}
    \Delta s_{t+1} &\sim f^{\Delta s_{t+1}}_\theta(\cdot|s^*_{t}, s)\\
    \mathrm{with}\quad f^{\Delta s_{t+1}}_\theta(\Delta s_{t+1}|s^*_{t}, s)&=\prod_{i=1}^n \prod_{k=1}^2 f^{\Delta v_{i,k}^{s_{t+1}}}_\theta(\Delta v_{i,k}^{s_{t+1}} \mid s^*_{t}, s)
\end{split}
\end{equation}

Moreover, since each $\Delta v_{i,k}^{s_{t+1}}$ is modeled as a continuous number, each $f^{\Delta v_{i,k}^{s_{t+1}}}_\theta(\cdot|s^*_{t},s)$ is then modeled as a Gaussian distribution:
\begin{equation}
f^{\Delta v_{i,k}^{s_{t+1}}}_\theta(\cdot|s^*_{t},s)=\mathcal{N}\big(\mu_\theta(\Delta v^{s_{t+1}}_{i,k} | s^*_{t}, s),\sigma_\theta(\Delta v^{s_{t+1}}_{i,k} | s^*_{t}, s)^2\big)
\end{equation}
where $f_\theta$ predicts a mean $\mu_\theta(\Delta v^{s_{t+1}}_{i,k} | s^*_{t}, s)$ and a standard deviation $\sigma_\theta(\Delta v^{s_{t+1}}_{i,k} | s^*_{t}, s)$ for each coordinate dimension. The mean determines the expected value of the coordinate offset, whereas the standard deviation controls the shape of the distribution.

\vspace{1mm}\noindent\textbf{Unsupervised training strategy.}\quad To train $f_\theta$ in an unsupervised manner, we first define its optimization objective. Given an original TSP instance $s$ and a base traditional heuristic tour constructor, $f_\theta$ refines $s$ auto-regressively, aiming to enable the base tour constructor to produce tours shorter than its original tours on $s$. A straight-forward optimization objective for such auto-regressive model is to maximize the total cumulative tour length reduction at the final modification iteration $T$ (i.e., $c_s(\pi^{s^*_0})-c_s(\pi^{s^*_{T}})$). However, since the number of iterations $T$ at inference can be arbitrarily chosen in our setting, training $f_\theta$ to maximize the final reduction for a fixed $T$ may lead to ineffective intermediate modifications and suboptimal early stopping. 
To accommodate this flexibility, we instead adopt a greedy optimization objective that independently maximizes the tour length reduction at each iteration $t$:
\begin{equation}
\label{eq:obj}
    J_t(\theta)=\mathbb{E}_{(s^*_{t},s)\sim D_t,\Delta s_{t+1} \sim f_\theta^{\Delta s_{t+1}}(\cdot|s^*_{t},s)} 
    \big[ c_s(\pi^{s^*_{t}}) - c_s(\pi^{s+ \Delta s_{t+1}}) \big]
\end{equation}
where $D_t$ is the dataset for iteration $t$. 
This ensures that each modification iteration can effectively reduce the tour length on $s$, making $f_\theta$ effective under varying iteration budgets.

However, $f_\theta$ cannot be directly trained to optimize the objective $J_t(\theta)$, since the tour $\pi^{s+ \Delta s_{t+1}}$ is computed by the traditional heuristic tour constructor, it is non-differentiable with respect to the parameters of $f_\theta$.
A common approach to train $f_\theta$ is to adopt REINFORCE-based \cite{williams1992simple} gradient update rule:
\begin{equation}
\label{eq:reinforce}
\begin{split}
\nabla_\theta J_t(\theta) &= 
\mathbb{E}_{(s^*_{t},s)\sim D_t, \Delta s_{t+1} \sim f^{\Delta s_{t+1}}_\theta(\cdot | s^*_{t},s)} \Big[ 
A(s^*_{t},s+ \Delta s_{t+1})\\ 
&\quad \cdot\nabla_\theta \log f^{\Delta s_{t+1}}_\theta(\Delta s_{t+1} | s^*_{t}, s) 
\Big] \\
\mathrm{with}\quad &A(s^*_{t}, s + \Delta s_{t+1}) = (c_s(\pi^{s^*_{t}}) - c_s(\pi^{s+\Delta s_{t+1}})) - b(s,t,t+1) 
\end{split}
\end{equation}
where $b(s,t,t+1)$ is a baseline function used to compute the advantage $A(s^*_{t}, s + \Delta s_{t+1})$ to reduce the variance of the gradients.

Intuitively, REINFORCE shapes the distribution by increasing the probability of sampling modified instances that lead to larger tour length reductions, while decreasing the probability of less effective ones. 
In this way, $f_\theta$ can be trained without ground-truth labels by auto-regressively sampling instance modifications from $f^{\Delta s_{t+1}}_\theta(\cdot|s^*_{t},s)$ for each instance $s$ in the training batch, and updating its parameters at each iteration based on the sampled instance modifications via Equation \ref{eq:reinforce}. Ideally, this can guide $f_\theta$ toward sampling high quality instance modifications and modified instances as training progresses. Note that, for simplicity, a fixed number of iterations $T$ is used during training.

Although this basic design provides a feasible way to train $f_\theta$ in an unsupervised manner, the search space of possible instance modifications is extremely large. Relying solely on this approach would require extensive sampling to obtain high-quality modifications, resulting in impractically long training times.
In the following sections, we introduce several key designs to enable $f_\theta$ to be trained effectively and practically without incurring costly training.

\subsection{Discretization of Coordinate Offset} 
\label{sec:discrete}
In Section~\ref{sec:basic}, the coordinate offset for each coordinate dimension of each node is modeled as a continuous variable and sampled from a Gaussian distribution whose mean and standard deviation are predicted by $f_\theta$. The mean specifies the expected offset of a coordinate, while the standard deviation controls the shape of the distribution around the mean. 

Although this continuous formulation offers flexibility in candidate offset values, it suffers from several limitations:
(1) it introduces a large continuous search space, requiring extensive sampling to discover high-quality modifications, which in turn slows down training;
(2) Gaussian distribution is unimodal where the probability mass is concentrated around the mean. A small standard deviation restricts exploration to a narrow region of search space around the mean, which risks trapping $f_\theta$ in local optima when the predicted mean is suboptimal, especially during early stage of training. A larger standard deviation encourages exploration of regions farther from the mean, but inevitably spreads the probability mass more broadly across the space, leading to overly random and unstable sampling. Thus, the current design lacks flexibility in exploration;
(3) sampling an exact target value (e.g., $0.00$) in a continuous setting is inherently difficult. This is because the probability of sampling any exact value from a Gaussian distribution is always zero, so each sample will deviate slightly from the target in an unstable manner. Moreover, stably estimating a precise mean is also particularly challenging for continuous value (i.e., regression). These factors lead to unstable sampling of coordinate offsets, and such instability can significantly alter the structure of the instances, making the sampling process difficult to control.

\vspace{1mm}\noindent\textbf{Improved modified instance sampling.}\quad To overcome these issues, we discretize the search space of coordinate offsets. Specifically, assuming that the valid range of a coordinate offset is $(-1,1)$, a continuous coordinate offset $\Delta v^{s_{t+1}}_{i,k}$ can be approximated as a weighted sum of discrete digits at different scales:
\begin{equation}
\label{eq:discret}
    \Delta v^{s_{t+1}}_{i,k}\approx g(\delta^{s_{t+1}}_{i,k},\Delta v^{s_{t+1}}_{i,k,1},...,\Delta v^{s_{t+1}}_{i,k,M}) = \delta^{s_{t+1}}_{i,k} \cdot \sum^M_{m=1}0.1^m \cdot \Delta v^{s_{t+1}}_{i,k,m}
\end{equation}
where $\delta^{s_{t+1}}_{i,k} \in \{-1,1\}$ represents the sign of $\Delta v^{s_{t+1}}_{i,k}$, each $\Delta v^{s_{t+1}}_{i,k,m}\in \{0,1,2,\dots,9\}$ is a discrete digit, and $M$ is a hyper-parameter that controls the discretization precision. 

Through this formulation, the search space of coordinate offset can be discretized into multiple sets of candidate digits at different scales, along with a set of candidate sign. For each coordinate dimension $k$ of each node $i$, $f_\theta$ can then be trained to predict a Categorical distribution (i.e., classification) for the sign $f^{\delta^{s_{t+1}}_{i,k}}_\theta(\cdot|s^*_{t},s)\in \mathbb{R}^2$, and for each discrete digit at different scale $f^{\Delta v^{s_{t+1}}_{i,k,m}}_\theta(\cdot|s^*_{t},s)\in \mathbb{R}^{10}$.
By controlling $M$, we can cover all possible offsets up to $M$-digit precision, where each digit set contains only $10$ possible values. The sampling of each coordinate offset $\Delta v^{s_{t+1}}_{i,k}$ in Equation \ref{eq:delta_sampling} can then be rewritten as follows:
\begin{equation}
\label{eq:offset_sampling}
\begin{split}
    \delta^{s_{t+1}}_{i,k} \sim f^{\delta^{s_{t+1}}_{i,k}}_\theta(\cdot|s^*_{t},s)&,\quad\Delta v^{s_{t+1}}_{i,k,m}\sim f^{\Delta v^{s_{t+1}}_{i,k,m}}_\theta(\cdot|s^*_{t},s), \quad m=1,\dots,M\\
    \Delta v^{s_{t+1}}_{i,k}&=g(\delta^{s_{t+1}}_{i,k},\Delta v^{s_{t+1}}_{i,k,1},...,\Delta v^{s_{t+1}}_{i,k,M})
\end{split}
\end{equation}
and the overall probability (density) of $\Delta v^{s_{t+1}}_{i,k}$ can be expressed as:
\begin{equation}
\label{eq:prob_offset}
    f^{\Delta v_{i,k}^{s_{t+1}}}_\theta(\Delta v_{i,k}^{s_{t+1}}|s^*_{t},s) = f^{\delta^{s_{t+1}}_{i,k}}_\theta(\delta^{s_{t+1}}_{i,k}|s^*_{t},s)\cdot\prod_{m=1}^M f^{\Delta v^{s_{t+1}}_{i,k,m}}_\theta(\Delta v^{s_{t+1}}_{i,k,m}|s^*_{t},s)
\end{equation}

This discretized formulation effectively addresses the limitations of the previous continuous formulation:
(1) the search space is reduced to a finite space; (2) the Categorical distribution enables flexible exploration beyond the mode of the distribution; (3) the Categorical distribution enables more stable sampling of precise coordinate offset values.

\subsection{Guiding $f_\theta$ with Self-imitation Learning}
\label{sec:imitation}
Although discretizing coordinate offsets substantially reduces the search space, unsupervised training of $f_\theta$ with REINFORCE still requires extensive sampling to converge.
In Section~\ref{sec:basic}, the basic unsupervised training strategy iteratively samples instance modifications $\Delta S_{t+1}$ using $f_\theta$, and updates its parameters based on the the tour length reductions of the sampled modifications via REINFORCE (Eq.~\ref{eq:reinforce}).
However, during early training, $f_\theta$ is unguided and tends to generate highly random and overly distorted instances, often producing low-quality tours with negative tour length reductions. Learning from such poor samples yields weak and unstable training signals for cold-start, requiring excessive sampling to discover effective modifications and making the training inefficient.

\vspace{1mm}\noindent\textbf{Improved unsupervised training strategy.}\quad To train $f_\theta$ more efficiently, a common strategy is to enhance REINFORCE with imitation learning.
Given a pair $(s^*_{t}, s)$, assuming that we have a high-quality expert demonstration ${\Delta s^{\mathrm{expert}}_{t+1}}$ (e.g., a near-optimal or optimal modification),
% that consists of expert coordinate offsets ${\Delta v_{i,k}^{s^\mathrm{expert}_{t+1}}}$ for each coordinate dimension $k$ of each node $i$,
$f_\theta$ is then trained to imitate ${\Delta s^{\mathrm{expert}}_{t+1}}$ by maximizing the likelihood of producing it, such that:
\begin{equation}
\label{eq:imitation_loss_basic}
\begin{split}
    J^\mathrm{IL}_t(\theta) &=\mathbb{E}_{(s^*_{t},s,\Delta s^\mathrm{expert}_{t+1})\sim D_t}\Big[\log f^{\Delta s_{t+1}}_\theta(\Delta s^\mathrm{expert}_{t+1}|s^*_{t}, s)\Big]\\
\end{split}
\end{equation}
where $f^{\Delta s_{t+1}}_\theta(\Delta s^\mathrm{expert}_{t+1}|s^*_{t}, s)$ can be computed by Eq.~\ref{eq:prob_basic} and Eq.~\ref{eq:prob_offset}.
The overall training objective $J^\mathrm{total}_t(\theta)$ can then be expressed as:
\begin{equation}
\label{eq:object_total}
    J^\mathrm{total}_t(\theta) = J^\mathrm{RL}_t(\theta) + \lambda \cdot J^\mathrm{IL}_t(\theta)
\end{equation}
where $J^\mathrm{RL}_t(\theta)$ is the optimization objective defined in Eq.~\ref{eq:obj} that maximizes the tour length reduction, and $\lambda$ is a hyperparameter that balances these two objectives.
Such expert guidance is particularly useful at the early stage of training, providing a cold-start signal that guides $f_\theta$ toward meaningful regions of the search space instead of relying solely on random low-quality samples, thereby reducing ineffective exploration and accelerating convergence.

However, directly obtaining near-optimal/optimal expert instance modifications is often prohibitively expensive in practice. A practical alternative is to adopt self-imitation learning, where we set $\Delta s^\mathrm{expert}_{t+1}$ to be the best modification found by $f_\theta$ itself, such that:
\begin{equation}
\begin{split}
    &\Delta s^\mathrm{expert}_{t+1}=\Delta s^*_{t+1} = s^*_{t+1}-s \\
    \mathrm{with}\quad & {\Delta v_{i,k}^{s^\mathrm{expert}_{t+1}}} = v_{i,k}^{s^*_{t+1}} - v_{i,k}^{s}, \quad i=1,\dots,n,\; k=1,2
\end{split} 
\end{equation}
Although $f_\theta$ tends to generate low-quality random modifications in early training, our formulation provides a natural safeguard: if all sampled instance modifications in $\Delta S_{t+1}$ are poor, $s^*_{t+1}$ can be simply set equal to $s^*_t$ (Eq.~\ref{eq:s_best}). In other words, $f_\theta$ can be initially trained to keep the current best modified instance $s^*_{t}$ unchanged for cold-start. 
This key design anchors the search space to slight deviations from the unchanged instance, preventing the model from generating overly random poor-quality modifications and enabling more efficient exploration. While this cold-start guidance is sub-optimal, it offers a simple yet effective regularization, and is reasonable, because modifications are only required for nodes in complex regions where the optimal sub-tour is difficult to find, rather than for every node in the instance. As training proceeds and better modifications are discovered, $f_\theta$ can then shift to imitate the best sampled modification. This self-imitation mechanism progressively guides $f_\theta$ toward higher-quality and more meaningful modifications. 
Moreover, it is worth noting that such a safeguard is not available to neural-based heuristic tour constructors that directly predict complete tours, they can only imitate their own poor-quality solutions in the early stage, which often leads to inefficient training.

Notably, $\Delta s^\mathrm{expert}_{t+1}$ in this self-imitation learning formulation are often neither near-optimal nor optimal, and the quality of $\Delta s^\mathrm{expert}_{t+1}$ could vary across different instances in a training batch (some modifications may yield significant improvements in tour length reduction, while others contribute little), particularly during the early stages of training. Therefore, we weight the imitation objective of each $\Delta s^\mathrm{expert}_{t+1}$ by its corresponding tour length reduction $c_s(\pi^{s^*_{t}}) - c_s(\pi^{s+\Delta s^\mathrm{expert}_{t+1}})$. Moreover, since the scale of tour-length reductions can differ across iterations $t$ (larger reductions are more likely at early iterations), we normalize the weights within each iteration. % by dividing the sum of reductions across all training instances. 
And since each initial guidance might yield no tour-length reduction (i.e., $c_s(\pi^{s^*_{t}}) - c_s(\pi^{s+\Delta s^\mathrm{expert}_{t+1}})=0$), its normalized weight would always be $0$ under the above scheme, causing the model to ignore this cold-start signal, thus we additional assign a fixed small weight to each $\Delta s^\mathrm{expert}_{t+1}$. The overall weight of each $\Delta s^\mathrm{expert}_{t+1}$ is:
\begin{equation}
    w^{\Delta s^\mathrm{expert}_{t+1}} =  \frac{c_s(\pi^{s^*_{t}}) - c_s(\pi^{s+\Delta s^\mathrm{expert}_{t+1}})}{\sum_{(s^*_{t},s,\Delta s^\mathrm{expert}_{t+1})\in D_t} c_s(\pi^{s^*_{t}}) - c_s(\pi^{s+\Delta s^\mathrm{expert}_{t+1}})} + w^\mathrm{fixed}
\end{equation}
and the objective $J^{\mathrm{IL}}_t(\theta)$ in Eq.~\ref{eq:imitation_loss_basic} can be rewritten as follows:
\begin{equation}
\label{eq:obj_imitation}
    J^{\mathrm{IL}}_t(\theta) \approx \sum_{(s^*_{t},s,\Delta s^\mathrm{expert}_{t+1})\in D_t}  w^{\Delta s^\mathrm{expert}_{t+1}} \cdot \log f^{\Delta s_{t+1}}_\theta(\Delta s^\mathrm{expert}_{t+1}|s^*_{t}, s)
\end{equation}

\subsection{Unifying Target Output Space across Nodes}
\label{sec:unification}
Given an original instance $s$ and a best modified instance $s^*_{t}$, the neural-based instance modifier is currently designed to produce distributions $f^{\Delta v_{i,k}^{s_{t+1}}}_\theta(\cdot|s^*_{t},s)$, from which refined coordinate offsets can be sampled and added to the corresponding original coordinate in $s$ (Eq.~\ref{eq:delta_sampling}).
However, the refined coordinate offset $\Delta v^{s_{t+1}}_{i,k}$ often depends on the previous coordinate offset $\Delta v^{s^*_t}_{i,k}$. Since $\Delta v^{s^*_t}_{i,k}$ can vary significantly not only across nodes within the same TSP instance but also across different instances, the optimal target output space for the refined offsets also differ substantially across nodes. This often leads to slow convergence and long training time before satisfactory performance can be achieved.

To illustrate this, consider a simple case where the target refined coordinate offsets for each node are identical to the current coordinate offset $\Delta v^{s^*_t}_{i,k}$ (i.e., we want to keep the modified instance unchanged). In this case, $f_\theta$ only needs to simply reproduce $\Delta v^{s^*_t}_{i,k}$ for $\Delta v^{s_{t+1}}_{i,k}$ for each coordinate dimension of each node. However, since the magnitude of $\Delta v^{s^*_t}_{i,k}$ vary significantly across different nodes and instances, $f_\theta$ still requires substantial training time even in this simple scenario. 
In more complex scenarios, $f_\theta$ must not only recover the current coordinate offsets but also learn meaningful refinement of the offsets that improve the tour quality on $s$, which further increases the training difficulty.

\vspace{1mm}\noindent\textbf{Improved modified instance sampling.}\quad To unify the target output space across different nodes, we propose to predict the distributions of coordinate offset refinement rather than the distributions of the refined new coordinate offset. In other words, $\Delta s_{t+1}$ is now set to $\Delta s_{t+1} = s_{t+1} - s^*_t$, rather than $\Delta s_{t+1} = s_{t+1} - s$, and $s_{t+1}$ is now sampled by sampling $\Delta s_{t+1}$ and adding it to the current modified instance $s^*_t$, such that:
\begin{equation}
\label{eq:unified_sampling}
\begin{split}
    \Delta v_{i,k}^{s_{t+1}} &\sim f^{\Delta v_{i,k}^{s_{t+1}}}_\theta(\cdot | s^*_{t}, s), 
    \quad i=1,\dots,n,\; k=1,2 \\
    s_{t+1} = s^*_t + \Delta s_{t+1},&\quad \Delta s_{t+1} = \{(\Delta v_{1,1}^{s_{t+1}},\Delta v_{1,2}^{s_{t+1}}),\dots,(\Delta v_{n,1}^{s_{t+1}},\Delta v_{n,2}^{s_{t+1}})\}\\
\end{split}
\end{equation}
Consider the previous example where no further modification is required: $f_\theta$ now only needs to learn that the coordinate offset refinement for all nodes are zero (i.e., $\Delta v^{s_{t+1}}_{i,k}=0$), and each node shares the same target output. This design effectively eliminates the large
variability in the target output space caused by differences in previous coordinate offset across nodes and instances, thereby significantly accelerating the training of $f_\theta$.

\vspace{1mm}\noindent\textbf{Improved unsupervised training strategy.}\quad Accordingly, the optimization objective in Eq.~\ref{eq:obj} can be rewritten as:
\begin{equation}
\label{eq:unified_reinforce_obj}
    J_t(\theta)=\mathbb{E}_{(s^*_{t},s)\sim D_t,\Delta s_{t+1} \sim f_\theta^{\Delta s_{t+1}}(\cdot|s^*_{t},s)} 
    \big[ c_s(\pi^{s^*_{t}}) - c_s(\pi^{s^*_t+ \Delta s_{t+1}}) \big]
\end{equation}
and the REINFORCE-based gradient update rule (Eq.~\ref{eq:reinforce}) can then be rewritten as:
\begin{equation}
\label{eq:unified_reinforce}
    \begin{split}
        \nabla_\theta J_t(\theta) &= 
\mathbb{E}_{(s^*_{t},s)\sim D_t, \Delta s_{t+1} \sim f^{\Delta s_{t+1}}_\theta(\cdot | s^*_{t},s)} \Big[ 
A(s^*_{t},s^*_{t}+ \Delta s_{t+1})\\ 
&\quad \cdot\nabla_\theta \log f^{\Delta s_{t+1}}_\theta(\Delta s_{t+1} | s^*_{t}, s) 
\Big] \\
    \end{split}
\end{equation}

Moreover, for the self-imitation learning, the expert demonstration $\Delta s^\mathrm{expert}_{t+1}$ can be computed as:
\begin{equation}
\begin{split}
    &\Delta s^\mathrm{expert}_{t+1}=\Delta s^*_{t+1} = s^*_{t+1}-s^*_{t} \\
    \mathrm{with}\quad & {\Delta v_{i,k}^{s^\mathrm{expert}_{t+1}}} = v_{i,k}^{s^*_{t+1}} - v_{i,k}^{s^*_{t}}, \quad i=1,\dots,n,\; k=1,2
\end{split} 
\end{equation}
and the weight of each $\Delta s^\mathrm{expert}_{t+1}$ can be represented as:
\begin{equation}
    w^{\Delta s^\mathrm{expert}_{t+1}} =  \frac{c_s(\pi^{s^*_{t}}) - c_s(\pi^{s^*_{t}+\Delta s^\mathrm{expert}_{t+1}})}{\sum_{(s^*_{t},s,\Delta s^\mathrm{expert}_{t+1})\in D_t} c_s(\pi^{s^*_{t}}) - c_s(\pi^{s^*_{t}+\Delta s^\mathrm{expert}_{t+1}})} + w^\mathrm{fixed}
\end{equation}

\vspace{1mm}\noindent\textbf{Due to the space limitation, the overall training pipeline of TSP-MDF can be found in Appendix~\ref{appendix:training_algo}. The architecture of neural-based instance modifier can be found in Appendix~\ref{appendix:network}.}

\begin{table*}[ht]
    \caption{Results on TSP-$500/1000/10000$. Each instance modification iteration of TSP-MDF samples $100$ modified instances.  ``AS'', ``MS'', and ``GS'' denote Active Search, Multi-step Sampling, and Gradient Search, respectively.  $^*$ indicates the baseline used to compute the performance gap. Results for exact/near-optimal TSP solvers are taken from ~\cite{qiu2022dimes, sun2023difusco, li2024fast}. For the training time, ``h'', ``m'', ``s'' denote hours, minutes, and seconds, respectively. ``OOM'' indicates that the method ran out of GPU memory ($24$ GB).}
    \vspace{-0.3cm}
    \centering
    \resizebox{\linewidth}{!}{\begin{tabular}{l|cccc|cccc|cccc}
        \toprule
        \multirow{2}*{Method}
        &\multicolumn{4}{c|}{TSP-500}&\multicolumn{4}{c|}{TSP-1000}&\multicolumn{4}{c}{TSP-10000}\\
        &Length $\downarrow$&Gap $\downarrow$&Inference Time $\downarrow$&Training Time $\downarrow$&Length $\downarrow$&Gap $\downarrow$&Inference Time $\downarrow$&Training Time $\downarrow$&Length $\downarrow$&Gap $\downarrow$&Inference Time $\downarrow$&Training Time $\downarrow$\\
        \midrule
        Concorde&$16.55^*$&$0.00\%$&$37$m$39$s&-&$23.12^*$&$0.00\%$&$6$h$39$m&-&N/A&N/A&N/A&-\\
        LKH-3&$16.55$&$0.00\%$&$46$m$40$s&-&$23.12$&$0.00\%$&$2$h$34$m&-&$71.77^*$&$0.00\%$&$8$h$48$m&-\\
        
        \midrule
        DIMES&$19.10$&$15.41\%$&$22$s&{$17$m$20$s}&$26.37$&$14.06\%$&$46$s&{$32$m$58$s}&$85.90$&$19.69\%$&$1$m$6$s&{$2$h$5$m$54$s}\\  
        DIMES+AS&$17.77$&$7.37\%$&$37$m$10$s&{$17$m$20$s}&$24.89$&$7.66\%$&$1$h$17$m$17$s&{$32$m$58$s}&$80.62$&$12.33\%$&$59$m$12$s&{$2$h$5$m$54$s}\\
        DeepACO&$18.93$&$14.38\%$&$16$m$25$s&$6$m$17$s&$27.47$&$18.81\%$&$32$m$22$s&$8$m$33$s&OOM&OOM&OOM&OOM\\
        Pointerformer&$17.14$&$3.56\%$&$53$s&{$547$h$47$m}&$24.80$&$7.27\%$&$6$m$35$s&{$2198$h$53$m}&OOM&OOM&OOM&OOM\\
        \midrule
        DIFUSCO&$17.60$&$6.34\%$&$10$m$46$s&{$129$h$10$m}&$25.09$&$8.52\%$&$45$m$17$s&{$295$h$50$m}&$93.70$&$30.56\%$&$1$h$25$m$15$s&OOM\\

        FastT2T&$17.68$&$6.82\%$&$25$s&{$260$h}&$24.88$&$7.61\%$&$1$m$45$s&OOM&$90.84$&$26.57$\%&$18$m$56$s&OOM\\
        FastT2T+MS+GS&$\textbf{16.71}$&$\textbf{0.97\%}$&$6$m$29$s&{$260$h}&OOM&OOM&OOM&OOM&OOM&OOM&OOM&OOM\\
        \midrule
        Nearest Insertion&$20.62$&$24.59\%$&$0.19$s&-&$28.95$&$25.21\%$&$0.72$s&-&$90.49$&$26.08\%$&$8.34$s&-\\
        TSP-MDF ($T=1$) + Nearest &$20.41$&$23.32\%$&$3$s&$2$m$01$s&$28.78$&$24.48\%$&$7$s&$5$m$47$s&$90.38$&$25.93\%$&$55$s&$1$h$37$m$31$s\\
        TSP-MDF ($T=30$) + Nearest &$18.94$&$14.44\%$&$1$m$11$s&$2$m$01$s&$27.09$&$17.17\%$&$3$m$6$s&$5$m$47$s&$88.86$&$23.81\%$&$24$m$52$s&$1$h$37$m$31$s\\
        \hdashline
        Farthest Insertion&$18.29$&$10.51\%$&$0.17$s&-&$25.73$&$11.29\%$&$0.65$s&-&$80.69$&$12.42\%$&$8.43$s&-\\
        TSP-MDF ($T=1$) + Farthest &$18.04$&$9.00\%$&$3$s&$1$m$55$s&$25.49$&$10.25\%$&$6$s&$5$m$21$s&$80.47$&$12.12\%$&$57$s&$1$h$35$m$34$s\\
        TSP-MDF ($T=30$) + Farthest &$17.26$&$4.29\%$&$1$m$6$s&$1$m$55$s&$\textbf{24.44}$&$\textbf{5.71\%}$&$2$m$34$s&$5$m$21$s&$\textbf{78.91}$&$\textbf{9.95\%}$&$25$m$56$s&$1$h$35$m$34$s\\
        \bottomrule
    \end{tabular}}
    \label{tab:overall_performance}
    \vspace{-0.3cm}
\end{table*}
\section{Experiments}
\subsection{Experimental Settings}
\subsubsection{Datasets}
To evaluate our TSP-MDF, we closely follow the standard procedure of previous works \cite{fu2021generalize,qiu2022dimes,sun2023difusco,li2023t2t,li2024fast,ye2024glop,jin2023pointerformer,ye2023deepaco} to conduct experiments on large-scale TSP-$n$ benchmark testsets of $500/1000/10000$ nodes generated by \cite{fu2021generalize}. Specifically, TSP-$500$ and TSP-$1000$ each contain $128$ 2-D Euclidean TSP test instances, and TSP-$10000$ contains $16$ 2-D Euclidean TSP test instances.
For each dataset of $n$ nodes, we generate the corresponding training instances on the fly by sampling $n$ nodes independently from a uniform distribution over the unit square $[0,1]^2$. \textbf{Due to the space limitation, an extra evaluation on the real-world benchmarks TSPLib~\cite{reinelt1991tsplib} can be found in Appendix~\ref{appendix:tsplib}.}

\subsubsection{Baselines}
To evaluate how our proposed TSP-MDF enhances traditional deterministic heuristic tour constructors, we adopt the following classical heuristic tour constructors as the base constructors: (1) \textbf{Nearest Insertion} determines node insertion order based on the nearest pairwise distance, and (2) \textbf{Farthest Insertion} determines node insertion order based on the farthest pairwise distance.

For a clear reference, we obtain ground-truth optimal tours using traditional exact/near-optimal TSP solvers, and we also compare the TSP-MDF-enhanced traditional heuristic tour constructors with neural-based heuristic tour constructors. \textbf{Due to the space limitation, the details of baselines can be found in Appendix~\ref{appendix:baseline}.}

\subsubsection{Evaluation Metrics}
We evaluate our TSP-MDF based on
the following metrics: (1) \textbf{Length}: the average length of the best approximated tours; (2) \textbf{Gap}($\%$): the relative performance gap in the length of the best approximated tour compared to the exact optimal tour or a reference baseline tour; (3) \textbf{Inference Time}: the total running time for solving all instances;
(4) \textbf{Training Time}: the total training time of the neural-based approach.

\vspace{1mm}\noindent\textbf{Due to the space limitation, the implementation details and computational resources can be found in Appendix~\ref{appendix:experiment_setting}}.

\subsection{Overall Performance}
In this section, we evaluate the performance of our TSP-MDF against baseline methods, both with and without the optional instance refinement stage. Note that, since the primary focus of our work is on the TSP tour construction, we disable the optional tour refinement stage to ensure a fair and clear comparison.
Table~\ref{tab:overall_performance} shows the performance of TSP-MDF that samples $100$ modified instances for each iteration, it is clear to see that the solution quality of both Farthest Insertion and Nearest Insertion can be enhanced by unlocking the guided-sampling capability through our TSP-MDF. In particular, with the optional instance refinement stage ($T=30$), by leveraging both our proposed parallel and sequential guided-sampling strategies, the enhanced Farthest Insertion can achieve solution quality comparable to neural-based heuristic tour constructors on TSP-$500$, and significantly outperforms all supervised and unsupervised neural-based heuristic tour constructors on larger instances TSP-$1000$ and TSP-$10000$. 
Although TSP-MDF introduces slightly longer inference time compared to original traditional heuristics, the overall inference time remains efficient compared to neural-based heuristic tour constructors. More importantly, TSP-MDF offers clear advantages in terms of practicality, as our neural-based instance modifier does not require any ground-truth supervision and has extremely short training time. Specifically, training the neural-based instance modifier on TSP-$500$ and TSP-$1000$ takes less than $10$ minutes on a single RTX 3090 GPU, while most of the neural-based heuristic tour constructors require several days of training. And the practical neural-based heuristic tour constructors (i.e., DIMES, DeepACO) that does not require costly training yields inferior performance compared to our TSP-MDF–enhanced Farthest Insertion. This highlights that our TSP-MDF is both effective, efficient and practical.

\begin{table}[t]
\caption{Evaluation on tours refined by 2-opt~\cite{croes1958method} local search.}
\vspace{-0.3cm}
    \centering
    \resizebox{0.9\linewidth}{!}{\begin{tabular}{c|c|c|c}
         \toprule
         Method&TSP-500&TSP-1000&TSP-10000\\
         \midrule
         Nearest Insertion + 2-opt&$18.82$&$26.35$&$84.35$\\
         TSP-MDF (T=1) + Nearest + 2-opt&$18.78$&$26.30$&$84.28$\\
         TSP-MDF (T=30) + Nearest  + 2-opt&$18.33$&$25.85$&$83.35$\\
         \midrule
         Farthest Insertion + 2-opt&$18.05$&$25.40$&$79.62$\\
         TSP-MDF (T=1) + Farthest  + 2-opt&$17.84$&$25.22$&$79.50$\\
         TSP-MDF (T=30) + Farthest  + 2-opt&$17.23$&$24.38$&$78.34$\\
         \bottomrule
    \end{tabular}}
    \label{tab:2opt}
    \vspace{-0.5cm}
\end{table}

\subsection{Can TSP-MDF really enhance the traditional deterministic heuristic tour constructors?}
\label{sec:ablation_rand}
Given a base traditional deterministic heuristic tour constructor, our TSP-MDF aims to enhance it by unlocking its guided-sampling capability through instance modification, thereby increasing its chance of escaping local optima to discover alternative better tours. To verify whether our TSP-MDF can really enhance the base tour constructor, we evaluate the following two key questions: Q(1) \textbf{``Can guided-sampling on modified instances effectively help the base tour constructor escape local optima?''}; Q(2) \textbf{``Is guided-sampling necessary, or would random sampling suffice?''}.

\vspace{1mm}\noindent\textbf{Q(1).}\quad A reasonable way to evaluate Q(1) is to iteratively refine both the tours constructed by the original base tour constructor and our TSP-MDF-enhanced tour constructor via local search (i.e., optional tour refinement stage). If both refined tours converge to similar tours, it suggests that the tour constructed by the TSP-MDF–enhanced tour constructor remains within the same or a nearby local optima as that of the original tour constructor, indicating that the base tour constructor has not been successfully enhanced.
To conduct this evaluation, we adopt the widely used 2-opt~\cite{croes1958method} local search to refine the tours. Notably, since 2-opt is computationally more expensive than the base tour constructor, for simplicity, we only refine the best tour found by our TSP-MDF–enhanced tour constructor instead of all sampled tours.

Table~\ref{tab:2opt} shows the length of refined tours on each dataset. 
It can be observed that the tours found by our TSP-MDF–enhanced tour constructors converge to significantly shorter refined tours than those produced by the original base tour constructors. This indicates that TSP-MDF effectively helps the base tour constructors escape local optima and guides them toward better regions in the solution space, thereby successfully enhancing their overall performance.

\vspace{1mm}\noindent\textbf{Q(2).}\quad To answer Q(2), we construct a variant model, TSP-MDF (random), which removes the neural-based instance modifier and instead samples discrete coordinate offsets randomly. 
To better demonstrate the behavior of the variant models, we report the average tour length of the best approximated tour both including/excluding the tour computed on the original $s$ as one of the candidate tours. Notably, including $s$ evaluates how the overall performance of the base tour constructor can be enhanced, whereas excluding $s$ evaluates the pure quality of the modified instances.
As shown in Table~\ref{tab:ablation}, without guided-sampling, the average tour length cannot be improved compared to the original tour constructor (i.e., all modified instances are worse than $s$). This is because the randomly sampled modified instances distort the original instance in an uncontrolled manner, misleading the base tour constructor to approximate poor-quality tours of extremely long lengths (e.g., larger than $4000$ on TSP-10000). Thus, random sampling fails to help the base tour constructor escape local optima, highlighting the necessity of our proposed guided-sampling strategy. 
Together, the results from Q(1) and Q(2) consistently demonstrate that TSP-MDF can effectively enhance traditional deterministic tour constructors. 

\vspace{1mm}\noindent\textbf{Due to the space limitation, more experimental results can be found in Appendix~\ref{appendix:experiment}.}
\vspace{-0.2cm}

\section{Conclusion}
We presented TSP-MDF, a novel instance modification framework for the Traveling Salesman Problem. By leveraging neural-based instance modifier for guided-sampling, TSP-MDF enables traditional deterministic heuristic tour constructors to effectively generate multiple high-quality candidate tours from modified instances, helping them escape local optima and improve solution quality. Meanwhile, the proposed formulation allows the instance modifier to be trained efficiently and practicality in an unsupervised manner. Extensive experiments on large-scale and real-world benchmarks demonstrate that TSP-MDF significantly enhances traditional heuristic tour constructors, achieving solution quality comparable to neural-based heuristic tour constructors, while being substantially more practical with much shorter training time.

\bibliographystyle{ACM-Reference-Format}
\bibliography{sample-base}

\newpage
\appendix
\section*{Appendix}

\section{Extended Related Works}
\label{appendix:related_works}
\subsection{Exact Methods for TSP}
Traditional exact solvers for the TSP~\cite{gurobi,applegate2006concorde,cplex2009v12} are mainly built upon linear programming and mixed integer linear programming frameworks. Among them, Concorde~\cite{applegate2006concorde} is one of the most famous state-of-the-art exact solvers
based on an advanced branch-and-cut algorithm. Despite their strong optimality guarantees, these exact solvers are computationally expensive and become impractical for large-scale TSP instances.

\subsection{Approximation Methods for TSP}
To improve efficiency, TSP solutions are commonly approximated using a two-stage framework: an initial tour construction stage that generates one or more candidate tours, followed by an optional tour refinement stage to improve solution quality.

\vspace{1mm}\noindent\textbf{Tour construction.}\quad
In the construction stage, classical heuristics such as Farthest and Nearest Insertion iteratively build tours by inserting nodes according to distance-based rules. These methods are computationally efficient and produce reasonable solutions, but are inherently deterministic and typically generate only a single tour per instance, limiting exploration.

Recently, neural-based heuristic tour constructors~\cite{qiu2022dimes,sun2023difusco,li2024fast,kwon2020pomo,joshi2020learning,jin2023pointerformer,li2023t2t,hottung2021efficient,fu2021generalize,kim2021learning,kool2018attention, ye2023deepaco,li2025diversity,goh2024hierarchical} have emerged as popular alternatives for constructing TSP tours. These methods aim to estimate distributions that highlight promising regions of the solution space, from which multiple candidate tours can be sampled and explored.
Early neural-based frameworks~\cite{qiu2022dimes,kwon2020pomo,joshi2020learning,jin2023pointerformer,hottung2021efficient,kim2021learning,kool2018attention} primarily focus on developing unsupervised reinforcement learning methods to sequentially sample and insert nodes into partial tours. While effective on small-scale instances, these approaches often struggle to scale to larger instances due to the combinatorial complexity of the solution space and the low sampling efficiency of reinforcement learning.
More recently, generative diffusion-based models~\cite{sun2023difusco,li2023t2t,li2024fast} trained with supervised learning have demonstrated promising performance on large-scale TSP instances. By generating multiple diverse and high-quality distributions for a given instance, these models are able to effectively capture the multimodal nature of the problem. However, they rely heavily on supervised training and require access to high-quality ground-truth solutions, which are computationally expensive to obtain in practice. Moreover, existing neural-based heuristic tour constructors typically need to be trained entirely from scratch and often incur substantial training costs before achieving satisfactory performance, even on relatively small instances, which limits their practicality in real-world settings.

\vspace{1mm}\noindent\textbf{Tour refinement.}\quad To refine the candidate tours constructed by heuristic tour constructors, classical local search methods, such as $2$-opt~\cite{croes1958method} and LKH-3~\cite{helsgaun2017extension}, iteratively refine the tour through edge exchange operations. In particular, LKH-3 can achieve near-optimal solution quality by leveraging powerful $k$-opt moves combined with extensive multi-start strategies over a large number of randomized initial tours; however, this typically incurs substantial computational cost.
More recently, learning-based approaches~\cite{ye2024glop,kim2021learning,fu2021generalize,ye2023deepaco} have been explored for the tour refinement stage. Specifically, Monte Carlo Tree Search (MCTS)~\cite{fu2021generalize} has been adopted for tour improvement by searching over the solution distribution estimated by neural-based heuristic tour constructors. Methods such as LCP~\cite{kim2021learning} and GLOP~\cite{ye2024glop} propose to decompose an initial tour into sub-tours and apply neural networks trained via reinforcement learning to refine these substructures. DeepACO~\cite{ye2023deepaco} introduces neural-guided solution perturbations to enhance the traditional local search. These learning-based approaches offer greater flexibility and have the potential to improve both the effectiveness and efficiency of tour refinement, but most of them require training additional neural networks.
Nevertheless, the effectiveness of all existing refinement methods remains highly dependent on the quality of the initial tours. When the initial tour is suboptimal, the refinement process may converge to another suboptimal tour within the same local basin and may require a large number of refinement steps, leading to high computational overhead compared to initial tour construction. Consequently, improving the quality of initial tours remains a critical and complementary problem for effective TSP approximation.

\section{Detailed Methods}
\subsection{Training Algorithm}
\label{appendix:training_algo}

The training pipeline of TSP-MDF is illustrated in Algorithm~\ref{algo:training}.

\begin{algorithm}[t]
    \caption{TSP-MDF Training Procedure}
    \label{algo:training}
    \renewcommand{\algorithmicrequire}{\textbf{Input:}}
    \begin{algorithmic}[1]
        \REQUIRE a neural instance modifier $f_\theta$, a traditional heuristic tour constructor $\mathrm{Heuristic}(\cdot)$, number of instance modification iterations $T$, number of training epochs $E$;
        \FOR{$e = 1$ to $E$}
            \STATE Randomly sample a batch of TSP instances $D$;
            \STATE Compute $\pi^{s} \leftarrow \mathrm{Heuristic}(s)$ for each $s \in D$;
            
            \STATE Initialize $s^*_0 \leftarrow s$ for each $s \in D$;

            \FOR{$t=0$ to $T-1$}
                \FOR{each $s \in D$}
                    
                    \STATE Sample a set of instance modifications $\Delta S_{t+1}$ from $f_\theta^{\Delta s_{t+1}}(\cdot|s^*_t,s)$ according to Eq.~\ref{eq:offset_sampling}, Eq.~\ref{eq:unified_sampling};
                    \STATE Obtain a set of modified instances 
                    $S_{t+1} \leftarrow \{\,s^*_t + \Delta s_{t+1} \mid \Delta s_{t+1} \in \Delta S_{t+1}\,\}$;
                    \STATE Compute $\pi^{s_{t+1}} \leftarrow \mathrm{Heuristic}(s_{t+1})$ for each $s_{t+1} \in S_{t+1}$;
                    
                    \STATE $s^*_{t+1} \leftarrow \arg\min_{s'\in\{s^*_{t}\}\cup S_{t+1}} c_s(\pi^{s'})$;
                    \STATE $\Delta s^{\mathrm{expert}}_{t+1} \leftarrow s^*_{t+1} - s^*_{t}$;   
                \ENDFOR
                \STATE Update $f_\theta$ by taking gradient step on $J_t^{\mathrm{total}}(\theta)$ using each $(s,s^*_t,\Delta S_{t+1},\Delta s^{\mathrm{expert}}_{t+1})$ according to Eq.~\ref{eq:object_total}, 
                Eq.~\ref{eq:obj_imitation}, 
                Eq.~\ref{eq:unified_reinforce_obj}, Eq.~\ref{eq:unified_reinforce};
            \ENDFOR
        \ENDFOR
   
    \end{algorithmic}
\end{algorithm}

\subsection{Model Architecture of Neural-based Instance Modifier}
\label{appendix:network}
For an original TSP instance $s$, in each iteration $t$, given a current best modified instance $s^*_t$ that consists of a set of modified node coordinates, the neural-based instance modifier $f_\theta$ takes $s$ and $s^*_t$ as inputs, and aims to further modify $s^*_t$ by predicting the distributions of coordinate offsets. For a fair comparison with neural-based heuristic tour constructors, we follow a similar two-step pipeline commonly used in existing neural-based heuristic tour constructors \cite{joshi2020learning,fu2021generalize,li2024fast,sun2023difusco,qiu2022dimes}, where we first jointly model $s$ and $s^*_t$ to construct a single graph $G_t$ that consists of a set of nodes and edges, then apply a graph neural network to $G_t$ to estimate the distributions of coordinate offsets for each node.

\vspace{1mm}\noindent \textbf{Graph construction.}\quad 
To efficiently model $s$ and $s^*_t$ jointly as a single graph at each iteration, we first construct a static base graph $G$ based on the original TSP instance $s$, and then dynamically incorporate information from the modified instance $s^*_t$ into $G$ to obtain $G_t$.

To construct the static base graph $G$, we initialize the feature of each node $i$ in $G$, denoted as $\bm{x}_i$, with its corresponding $v^s_i=(v^s_{i,1},v^s_{i,2})$ in $s$.
We then construct edges by connecting each node to a fixed number of its nearest neighbors according to their Euclidean distances in $s$, which prevents quadratic edge growth in large graphs. The edge feature $\bm{e}_{ij}$ for each edge $(i,j)$ in $G$ is set to the corresponding Euclidean distance $\|v^s_{i}-v^s_{j}\|_2$.

At each iteration $t$, we then update $G$ to $G_t$ by concatenating each node feature $x_i$ in $G$ with the information derived from its corresponding modified coordinates $v^{s^*_t}_i=(v^{s^*_t}_{i,1},v^{s^*_t}_{i,2})$ in $s^*_t$.
Specifically, for each node $i$, we first compute its total coordinate offsets $v^{s^*_t}_{i}-v^s_{i}=(v^{s^*_t}_{i,1}-v^s_{i,1},v^{s^*_t}_{i,2}-v^s_{i,2})$, then discretize each input offset $v^{s^*_t}_{i,k}-v^s_{i,k}$ into $M+1$ discrete components via $g^{-1}(v^{s^*_t}_{i,k}-v^s_{i,k})$ (Eq.~\ref{eq:discret}) to capture the variations in the input offsets more precisely. Finally, each discrete component is transformed into a one-hot vector, and the feature of node $i$ in $G_t$, denoted as $\bm{x}^t_{i}$, is then obtained by concatenating all these one-hot vectors with its corresponding node feature $x_i$ in $G$. For simplicity and computational efficiency, $G_t$ preserves the edge set of $G$, with each edge feature denoted as $\bm{e}^t_{ij}$.

\vspace{1mm}\noindent \textbf{Network architecture.}\quad Given the constructed graph $G_t$, we follow previous neural-based heuristic tour constructors \cite{joshi2020learning,li2024fast,sun2023difusco,qiu2022dimes} on the choice of the network architecture, where an anisotropic graph neural networks (AGNN) is adopted as the backbone of $f_\theta$ to simultaneously compute the embeddings of nodes and edges in $G_t$ through an edge-gating mechanism \cite{bresson2018experimental}. 

Let $\bm{h}^l_{i}$ and $\bm{h}^l_{ij}$ denote the embeddings of node $i$ and edge $(i,j)$ at layer $l$ of AGNN, respectively. The initial embeddings $\bm{h}^0_i$ and $\bm{h}^0_{ij}$ are obtained by applying nonlinear transformations to the corresponding node features $\bm{x}^t_i$ and edge features $\bm{e}^t_{ij}$ in $G_t$, respectively:
\begin{equation}
    \bm{h}^0_i = \alpha (\bm{\theta}_x \bm{x}^t_i),\quad \bm{h}^0_{ij}=\alpha(\bm{\theta}_e \bm{e}^t_{ij})
\end{equation}
where $\alpha$ denotes the SiLU activation\cite{elfwing2018sigmoid}, and $\bm{\theta}_x$, $\bm{\theta}_e$ are learnable parameters.

At each layer of AGNN, the embeddings of nodes and edges are updated following an anisotropic message passing scheme:
\begin{equation}
\begin{split}
    \bm{h}^l_i &= \bm{h}^{l-1}_{i} + \alpha(\mathrm{BN}(\bm{\theta}^l_1 h^{l-1}_i + \mathcal{A}_{j \in \mathcal{N}(i)}(\sigma(\bm{h}^{l-1}_{ij})\odot \bm{\theta}^l_2 \bm{h}^{l-1}_j)))\\
    \bm{h}^l_{ij} &=  \bm{h}^{l-1}_{ij} + \alpha(\mathrm{BN}(\bm{\theta}^l_3 \bm{h}^{l-1}_{ij} + \bm{\theta}^l_4 \bm{h}^{l-1}_i + \bm{\theta}^l_5 \bm{h}^{l-1}_j))
\end{split}
\end{equation}
where $\bm{\theta}^l_1$, $\bm{\theta}^l_2$, $\bm{\theta}^l_3$, $\bm{\theta}^l_4$, $\bm{\theta}^l_5$ are learnable parameters at layer $l$, $\mathrm{BN}$ denotes the Batch Normalization \cite{ioffe2015batch}, $\mathcal{A}$ represents the aggregation function (e.g., mean pooling), $\mathcal{N}(i)$ denotes the neighborhoods of node $i$, $\sigma$ is the sigmoid function, and $\odot$ is the Hadamard product. 

After a $L$-layer AGNN, a MLP is applied to each node embedding $\bm{h}^L_i$ to predict a vector of dimension $2(10M+2)$, which is then split into $2(M+1)$ probability distributions corresponding to each discretized component of the coordinate offset. %(i.e., $f^{\delta^{s_{t+1}}_{i,k}}_\theta(\cdot|s^*_{t},s)$, $f^{\Delta v^{s_{t+1}}_{i,k,m}}_\theta(\cdot|s^*_{t},s)$, see Eq.~\ref{eq:offset_sampling}).

\section{Experimental Settings}
\label{appendix:experiment_setting}
\subsection{Baselines}
\label{appendix:baseline}
To evaluate how our proposed TSP-MDF enhances traditional deterministic heuristic tour constructors, we adopt the following classical heuristic tour constructors as the base constructors: (1) \textbf{Nearest Insertion} determines node insertion order based on the nearest pairwise distance, and (2) \textbf{Farthest Insertion} determines node insertion order based on the farthest pairwise distance.

For a clear reference, we obtain ground-truth solutions using the following traditional exact/near-optimal TSP solvers:
(1) \textbf{Concorde}~\cite{applegate2006concorde} is an exact TSP solver, and (2) \textbf{LKH-3}~\cite{helsgaun2017extension} is a near-optimal solver that iteratively refines multiple random initial tours via $k$-opt operations.

We further compare the TSP-MDF-enhanced tour constructors with both supervised and unsupervised neural-based heuristic tour constructors.
Unsupervised neural-based heuristic tour constructors include: (1) \textbf{DIMES}~\cite{qiu2022dimes}, (2) \textbf{DeepACO}~\cite{ye2023deepaco}, and (3) \textbf{Pointerformer}~\cite{jin2023pointerformer}, which are recent SOTA reinforcement learning–based approaches.
Supervised neural-based heuristic tour constructors include:
(1) \textbf{DIFUSCO}~\cite{sun2023difusco} 
and (2) \textbf{FastT2T}~\cite{li2024fast}, which are recent SOTA supervised methods based on generative diffusion model. 

Similar to the optional instance refinement stage of our TSP-MDF, some neural-based heuristic tour constructors also support optional sequential tour sampling strategies during inference, i.e., \textbf{DIMES} can employ \textbf{Active Search}~\cite{qiu2022dimes,hottung2021efficient}, while \textbf{FastT2T} can adopt \textbf{Multi-step Sampling}~\cite{li2024fast} and \textbf{Gradient Search}~\cite{li2024fast}.

%Notably, some of the latest neural-based heuristic tour constructors~\cite{goh2024hierarchical,li2025diversity} are not included in our comparison, as they are either not publicly available, or were originally evaluated only on small-scale instances and cannot be feasibly trained on our large-scale datasets using a single NVIDIA RTX 3090 (24GB) GPU.

\begin{figure*}[ht]
    \centering
    \begin{subfigure}[t]{0.32\linewidth}
        \centering
        \includegraphics[width=\linewidth]{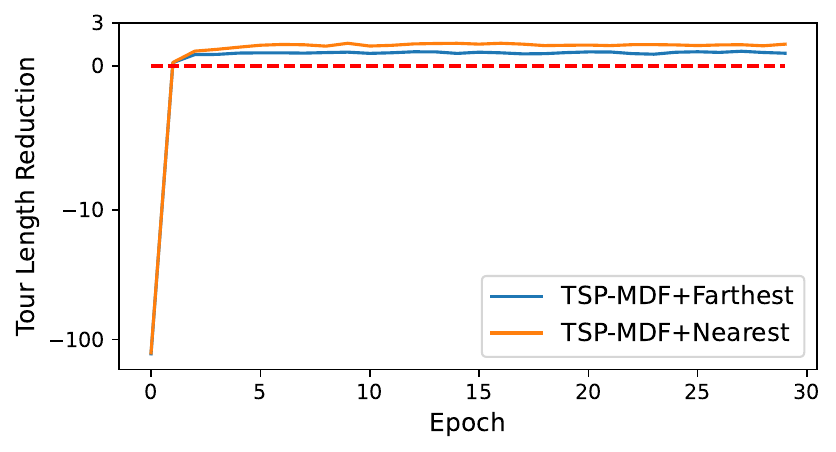}
        \caption{TSP-$500$}
        \label{fig:a}
    \end{subfigure}
    \hfill
    \begin{subfigure}[t]{0.32\linewidth}
        \centering
        \includegraphics[width=\linewidth]{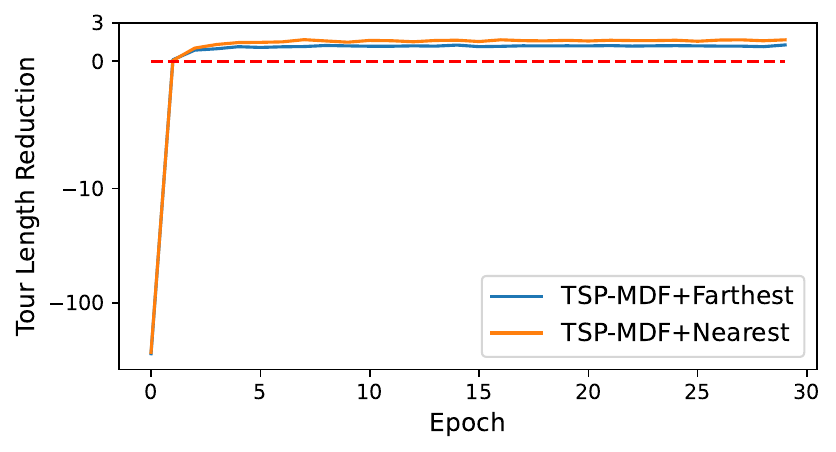}
        \caption{TSP-$1000$}
        \label{fig:b}
    \end{subfigure}
    \hfill
    \begin{subfigure}[t]{0.32\linewidth}
        \centering
        \includegraphics[width=\linewidth]{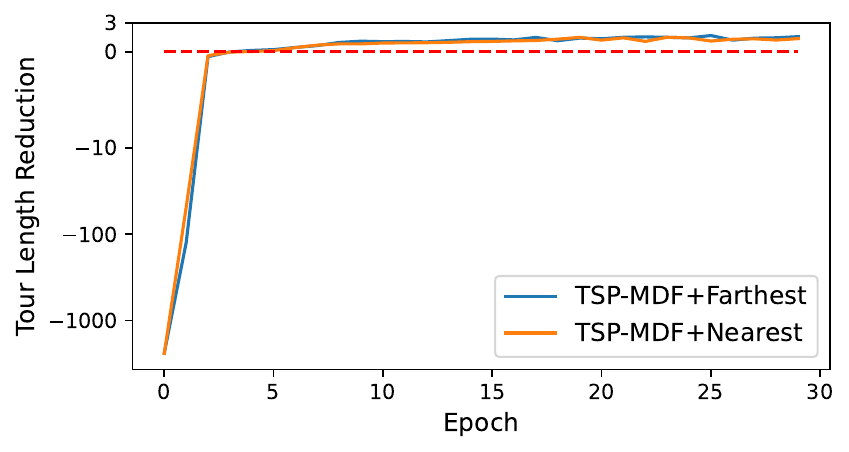}
        \caption{TSP-$10000$}
        \label{fig:c}
    \end{subfigure}
    \caption{Average tour length reduction of TSP-MDF-enhanced traditional heuristic tour constructors on training TSP instances.}
    \label{fig:training_curve}
    \vspace{-0.2cm}
\end{figure*}

\subsection{Implementation Details}
\noindent\textbf{Training.}\quad The neural-based instance modifier is trained using the AdamW optimizer~\cite{loshchilov2017decoupled} with a learning rate of $0.001$. We set the total number of training epochs $E=30$. In each training epoch, we sample $16$ original TSP instances (i.e., batch size) for the $500$-node and $1000$-node datasets, and $4$ instances for the $10,000$-node dataset. The total number of modification iterations per training epoch is set to $T=30$, and $|S_t|=50$ modified instances are sampled in each iteration for each original instance, with the discretization precision of sampling is set to $M=4$ for the $500$-node and $1000$-node datasets, and $M=6$ for the $10,000$-node dataset. For the training objective, we simply set $\lambda=1$, $w^\mathrm{fixed}=0.01$. 

\vspace{1mm}\noindent\textbf{Model architecture.}\quad For graph construction, each node is connected to its $50$ nearest neighbors. For the network parameters of neural-based instance modifier, we follow previous neural-based heuristic tour constructor~\cite{qiu2022dimes} to adopt a $12$-layer AGNN with a hidden dimension of $32$.

\subsection{Computational Resources}
All experiments are conducted on a single Nvidia GeForce RTX 3090 (24GB) GPU and an Intel i9-12900K 24-core CPU with 128GB RAM. In our TSP-MDF framework, the base traditional heuristic tour constructors are implemented in C++ and executed on the CPU, while the neural-based instance modifier is implemented in PyTorch and runs on the GPU.

\section{Additional Experimental Results}
\label{appendix:experiment}

\begin{table}[t]
\caption{Decomposition of training time for each instance modification iteration.}
\vspace{-0.3cm}
    \centering
    \resizebox{\linewidth}{!}{\begin{tabular}{c|c|c|c|c}
         \toprule
         Method&Component&TSP-500&TSP-1000&TSP-10000\\
         \midrule
         \multirow{3}{*}{TSP-MDF+Nearest}&Neural-based instance modifier&$0.066$s&$0.11$s&$0.30$s\\
         &Nearest Insertion&$0.064$s&$0.27$s&$6.20$s\\
         &Total&$0.13$s&$0.38$s&$6.5$s\\
         \midrule
         \multirow{3}{*}{TSP-MDF+Farthest}&Neural-based instance modifier&$0.062$s&$0.12$s&$0.24$s \\
         &Farthest Insertion&$0.058$s&$0.24$s&$6.13$s\\
         &Total&$0.12$s&$0.36$s&$6.37$s\\
         \bottomrule
    \end{tabular}}
    
    \label{tab:training_time}
    \vspace{-0.5cm}
\end{table}

\subsection{Practicability Analysis: Training Efficiency}
In this section, we analyze the practicability of our TSP-MDF in terms of the training efficiency, which can be decomposed into two components: the training time and the training dynamic.

\vspace{1mm}\noindent\textbf{Training time.}\quad Table~\ref{tab:training_time} presents the breakdown of training time for each instance modification iteration (the inference time follows the similar decomposition). For the dataset with $500$ nodes (i.e., TSP-$500$), the training time of the neural-based instance modifier (including both forward pass and backward propagation) is similar to the running time of the base heuristic tour constructor. As the instance size increases, the running time of the base tour constructor grows rapidly and dominates the overall training time, whereas the training time of our neural-based instance modifier remains efficient even for TSP-$10000$, which demonstrates the efficiency of neural-based instance modifier. Notably, the base heuristic tour constructor approximates tours on multiple sampled modified instances in parallel on the CPU, thus this process can be further accelerated by running it on a CPU with more cores.

\vspace{1mm}\noindent\textbf{Training dynamic.}\quad Figure~\ref{fig:training_curve} shows the average tour length reduction on the training TSP instances achieved by TSP-MDF-enhanced traditional heuristic tour constructors. Notably, the tour length reduction for each base tour constructor is computed by the difference between the length of tour originally approximated on $s$ and the length of the best tour found among all modified instances across all iterations (excluding $s$) sampled by the neural-based instance modifier. It can be observed that the neural-based instance modifier initially generates instance modifications of poor quality that mislead the base tour constructor to approximate extremely long tours, resulting in negative tour length reductions. But benefiting from our design, within only a few epochs, the neural-based instance modifier can quickly correct the direction and learn to generate meaningful instance modifications that result in positive tour reductions. 

Taken together, both the computational efficiency and learning capability of our neural-based instance modifier allow it to be trained efficiently, thus making TSP-MDF a practical framework.

\begin{table*}[t]
\caption{Effect of parallel sampling $|S_t|$ and sequential steps $T$ on tour length/inference time.}
\vspace{-0.3cm}
\centering

% ---------- 第一行 ----------
\begin{subtable}{0.49\textwidth}
\centering
\caption{TSP-500 Nearest Insertion}
\resizebox{\linewidth}{!}{\begin{tabular}{c|cccccc}
$|S_t|$ $\backslash$ $T$ & $1$ & $10$ & $20$ & $30$ & $40$ & $50$ \\
\hline
$10$  &$20.54$/$1$s &$20.06$/$10$s &$19.78$/$19$s &$19.61$/$29$s &$19.48$/$39$s &$19.38$/$48$s \\
$50$  &$20.46$/$2$s &$19.74$/$16$s &$19.37$/$31$s &$19.16$/$46$s &$19.02$/$60$s &$18.92$/$76$s \\
$100$ &$20.41$/$3$s &$19.63$/$26$s &$19.23$/$52$s &$18.94$/$71$s &$18.86$/$103$s &$18.77$/$128$s \\
$300$ &$20.39$/$6$s &$19.47$/$56$s &$19.06$/$111$s &$18.84$/$166$s &$18.69$/$221$s &$18.61$/$277$s \\
$500$ &$20.36$/$8$s &$19.41$/$83$s &$18.99$/$166$s &$18.76$/$250$s &$18.63$/$333$s &$18.53$/$416$s \\
\end{tabular}}
\end{subtable}
\hfill
\begin{subtable}{0.49\textwidth}
\centering
\caption{TSP-500 Farthest Insertion}
\resizebox{\linewidth}{!}{\begin{tabular}{c|cccccc}
$|S_t|$ $\backslash$ $T$ & $1$ & $10$ & $20$ & $30$ & $40$ & $50$ \\
\hline
$10$  &$18.19$/$1$s &$17.80$/$8$s &$17.64$/$16$s &$17.56$/$25$s &$17.50$/$33$s &$17.46$/$41$s \\
$50$  & $18.07$/$2$s&$17.54$/$15$s &$17.38$/$30$s &$17.31$/$44$s &$17.26$/$59$s &$17.24$/$74$s \\
$100$ &$18.04$/$3$s &$17.47$/$22$s &$17.31$/$42$s &$17.26$/$66$s &$17.21$/$84$s &$17.20$/$106$s \\
$300$ &$17.98$/$5$s &$17.39$/$53$s &$17.25$/$106$s &$17.19$/$160$s &$17.17$/$213$s &$17.16$/$266$s \\
$500$ &$17.96$/$7$s &$17.35$/$72$s &$17.21$/$144$s &$17.17$/$216$s &$17.15$/$289$s &$17.14$/$361$s \\
\end{tabular}}
\end{subtable}
\vspace{0.5em} 

\begin{subtable}{0.49\textwidth}
\centering
\caption{TSP-1000 Nearest Insertion}
\resizebox{\linewidth}{!}{\begin{tabular}{c|cccccc}
$|S_t|$ $\backslash$ $T$ & $1$ & $10$ & $20$ & $30$ & $40$ & $50$ \\
\hline
$10$  &$28.87$/$3$s &$28.44$/$25$s &$28.13$/$50$s &$27.90$/$74$s &$27.73$/$99$s &$27.59$/$124$s \\
$50$  &$28.81$/$5$s &$28.10$/$37$s &$27.65$/$72$s &$27.34$/$109$s &$27.11$/$144$s &$26.94$/$180$s \\
$100$ &$28.78$/$7$s &$27.99$/$56$s &$27.49$/$113$s &$27.09$/$186$s &$26.91$/$226$s &$26.72$/$282$s \\
$300$ &$28.75$/$14$s &$27.81$/$139$s &$27.24$/$278$s &$26.88$/$416$s &$26.62$/$555$s &$26.43$/$693$s \\
$500$ &$28.73$/$22$s &$27.75$/$219$s &$27.17$/$437$s &$26.80$/$655$s &$26.54$/$873$s &$26.34$/$1090$s \\
\end{tabular}}
\end{subtable}
\hfill
\begin{subtable}{0.49\textwidth}
\centering
\caption{TSP-1000 Farthest Insertion}
\resizebox{\linewidth}{!}{\begin{tabular}{c|cccccc}
$|S_t|$ $\backslash$ $T$ & $1$ & $10$ & $20$ & $30$ & $40$ & $50$ \\
\hline
$10$  &$25.63$/$3$s &$25.22$/$21$s &$25.01$/$41$s &$24.89$/$62$s &$24.71$/$82$s &$24.73$/$102$s \\
$50$  &$25.53$/$4$s &$24.95$/$34$s &$24.70$/$69$s &$24.57$/$103$s &$24.48$/$137$s &$24.42$/$170$s \\
$100$ &$25.49$/$6$s &$24.86$/$53$s &$24.59$/$106$s &$24.44$/$154$s &$24.37$/$211$s &$24.32$/$263$s \\
$300$ &$25.44$/$13$s &$24.72$/$127$s &$24.46$/$253$s &$24.32$/$378$s &$24.24$/$505$s &$24.19$/$632$s \\
$500$ &$25.41$/$25$s &$24.66$/$247$s &$24.39$/$494$s &$24.26$/$740$s &$24.19$/$986$s &$24.15$/$1233$s \\
\end{tabular}}
\end{subtable}
\vspace{0.5em}

\begin{subtable}{0.49\textwidth}
\centering
\caption{TSP-10000 Nearest Insertion}
\resizebox{\linewidth}{!}{\begin{tabular}{c|cccccc}
$|S_t|$ $\backslash$ $T$ & $1$ & $10$ & $20$ & $30$ & $40$ & $50$ \\
\hline
$10$  &$90.44$/$14$s &$90.09$/$99$s &$89.78$/$194$s &$89.58$/$290$s &$89.41$/$384$s &$89.27$/$480$s \\
$50$  &$90.40$/$33$s &$89.83$/$291$s &$89.38$/$578$s &$89.03$/$863$s &$88.76$/$1151$s &$88.55$/$1438$s \\
$100$ & $90.38$/$55$s &$89.74$/$706$s &$89.24$/$1396$s &$88.86$/$1492$s &$88.53$/$2794$s &$88.26$/$3493$s \\
$300$ &$90.36$/$213$s &$89.59$/$2100$s &$89.00$/$4231$s &$88.56$/$6381$s &$88.22$/$8540$s &$87.90$/$10691$s \\
$500$ &$90.37$/$367$s &$89.57$/$3580$s &$88.94$/$7111$s &$88.47$/$10721$s &$88.08$/$14309$s &$87.76$/$17881$s \\
\end{tabular}}
\end{subtable}
\hfill
\begin{subtable}{0.49\textwidth}
\centering
\caption{TSP-10000 Farthest Insertion}
\resizebox{\linewidth}{!}{\begin{tabular}{c|cccccc}
$|S_t|$ $\backslash$ $T$ & $1$ & $10$ & $20$ & $30$ & $40$ & $50$ \\
\hline
$10$  &$80.62$/$14$s &$80.21$/$103$s &$79.97$/$203$s &$79.77$/$303$s &$79.62$/$403$s &$79.48$/$503$s \\
$50$  &$80.53$/$35$s &$79.89$/$303$s &$79.50$/$597$s &$79.23$/$893$s &$79.01$/$1188$s &$78.84$/$1483$s \\
$100$ &$80.47$/$57$s &$79.79$/$520$s &$79.39$/$1035$s &$78.91$/$1556$s &$78.84$/$2068.98$s &$78.66$/$2588$s \\
$300$ &$80.48$/$198$s &$79.66$/$1978$s &$79.17$/$3988$s &$78.82$/$5982$s &$78.56$/$7988$s &$78.34$/$9982$s \\
$500$ &$80.45$/$234$s &$79.55$/$2327$s &$79.03$/$4648$s &$78.69$/$6962$s &$78.42$/$9270$s&$78.19$/$11581$s \\
\end{tabular}}
\end{subtable}

\label{tab:parallel_sequential}

\end{table*}

\subsection{Parallel vs. Sequential Guided-sampling}
Given a base traditional heuristic tour constructor, our TSP-MDF enhances it by leveraging both parallel and sequential guided sampling. Table~\ref{tab:parallel_sequential} investigates the effects of different numbers of parallel samples $|S_t|$ and sequential sampling steps $T$ on both solution quality and efficiency.
First, it is evident that increasing either parallel or sequential sampling generally leads to shorter tour lengths produced by the base heuristic tour constructor, with linearly increased inference time. However, when relying solely on parallel sampling without sequential refinement (i.e., $T=1$), the improvement in tour length is rather limited when $|S_t|>100$. By incorporating our proposed sequential sampling strategy, the tour length can be reduced much more rapidly. This is because obtaining high-quality modifications in a single shot is challenging; sequential sampling effectively decomposes this complex task into multiple simpler steps. Interestingly, we observe that certain combinations with more sequential sampling steps and fewer parallel samples dominate those with fewer sequential steps but more parallel samples, in terms of both tour length and inference time. For example, the configuration $(|S_t|=100, T=30)$ consistently outperforms $(|S_t|=300, T=20)$ across all datasets and for both heuristic tour constructors, achieving shorter tours while requiring less inference time. Notably, although TSP-MDF is trained using a fixed number of sequential steps ($T=30$), it remains effective during inference with an arbitrarily number of steps, including both fewer ($T<30$) and more ($T>30$) sequential steps, which demonstrates the flexibility of our sequential sampling.

\begin{figure*}[tbp]
\centering
    \begin{minipage}{\textwidth}

    \begin{subfigure}{0.32\textwidth}
        \centering
        \includegraphics[width=\textwidth]{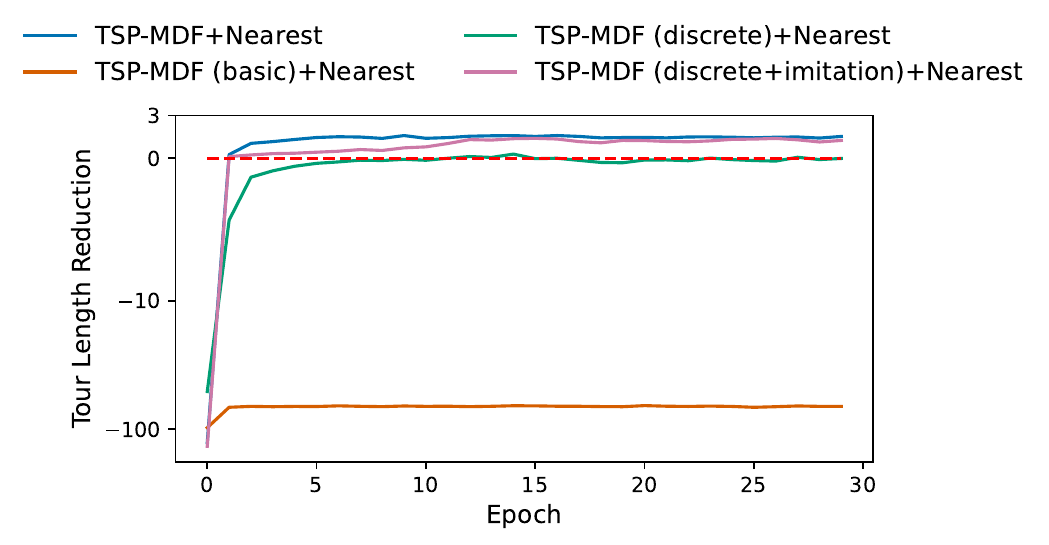}
        \caption{TSP-500 Nearest Insertion}
    \end{subfigure}
    \hfill
    \begin{subfigure}{0.32\textwidth}
        \centering
        \includegraphics[width=\textwidth]{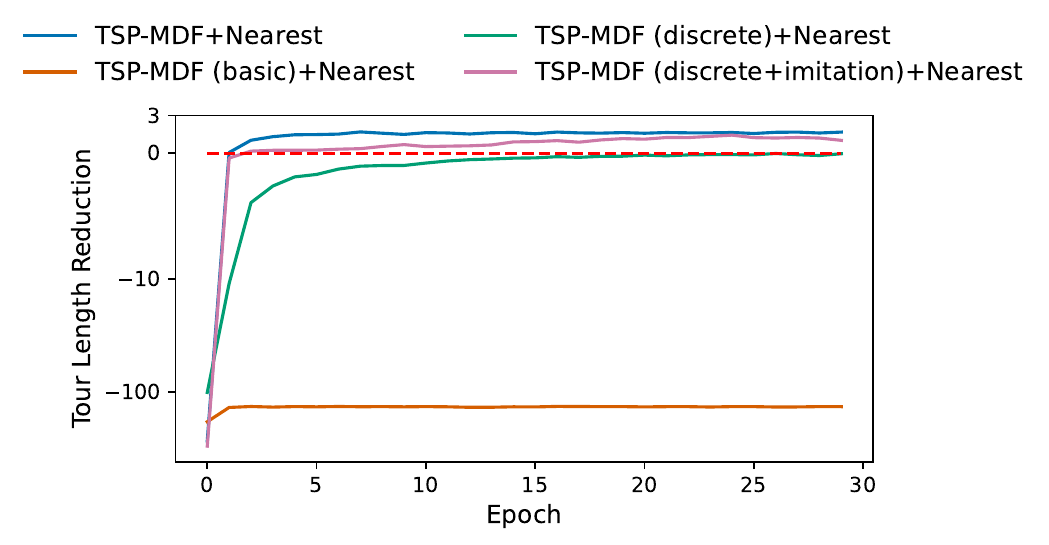}
        \caption{TSP-1000 Nearest Insertion} 
    \end{subfigure}
    \hfill
    \begin{subfigure}{0.32\textwidth}
        \centering
        \includegraphics[width=\textwidth]{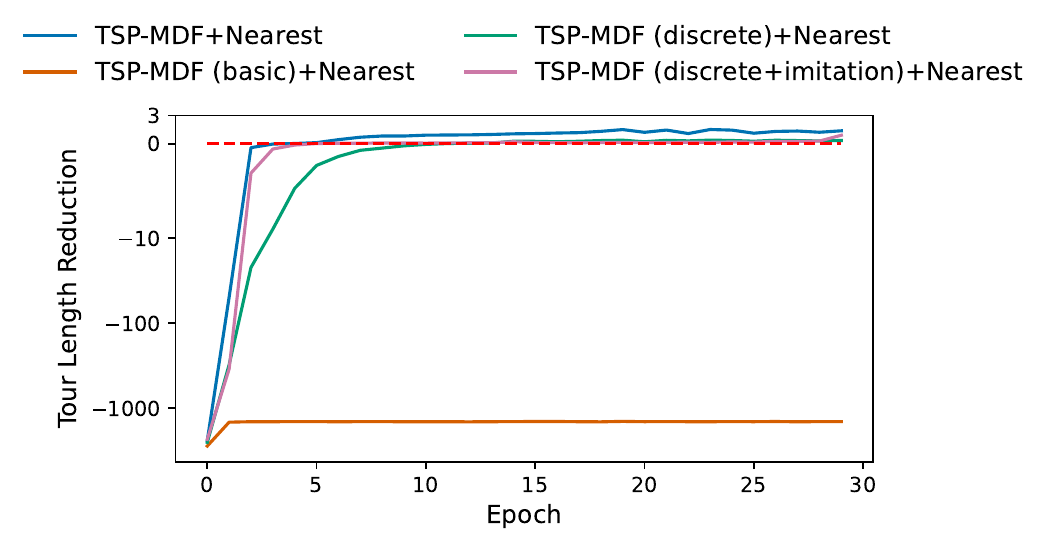}
        \caption{TSP-10000 Nearest Insertion} 
    \end{subfigure}

    \vskip\baselineskip
     \begin{subfigure}{0.32\textwidth}
        \centering
        \includegraphics[width=\textwidth]{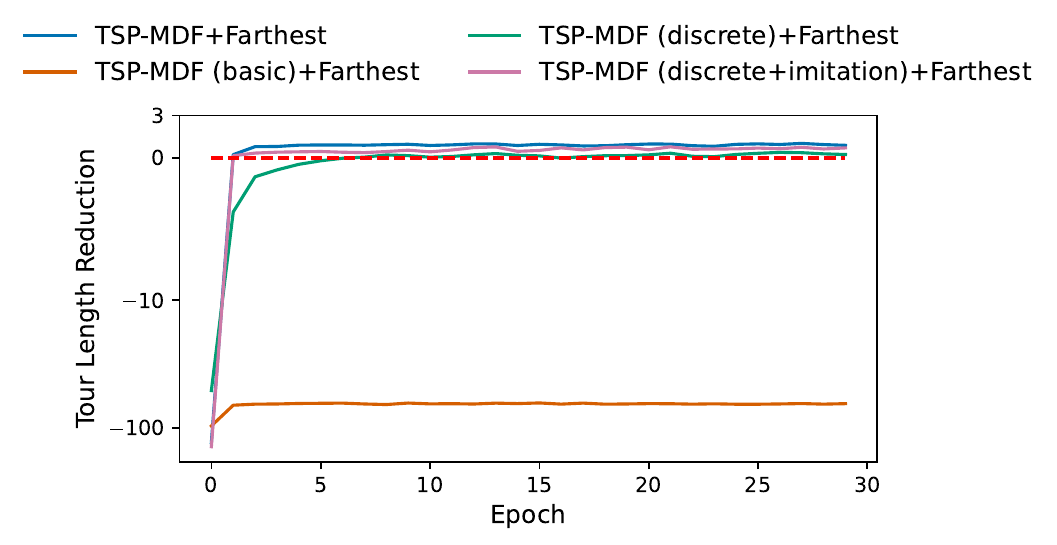}
        \caption{TSP-500 Farthest Insertion} 
    \end{subfigure}
    \hfill
    \begin{subfigure}{0.32\textwidth}
        \centering
        \includegraphics[width=\textwidth]{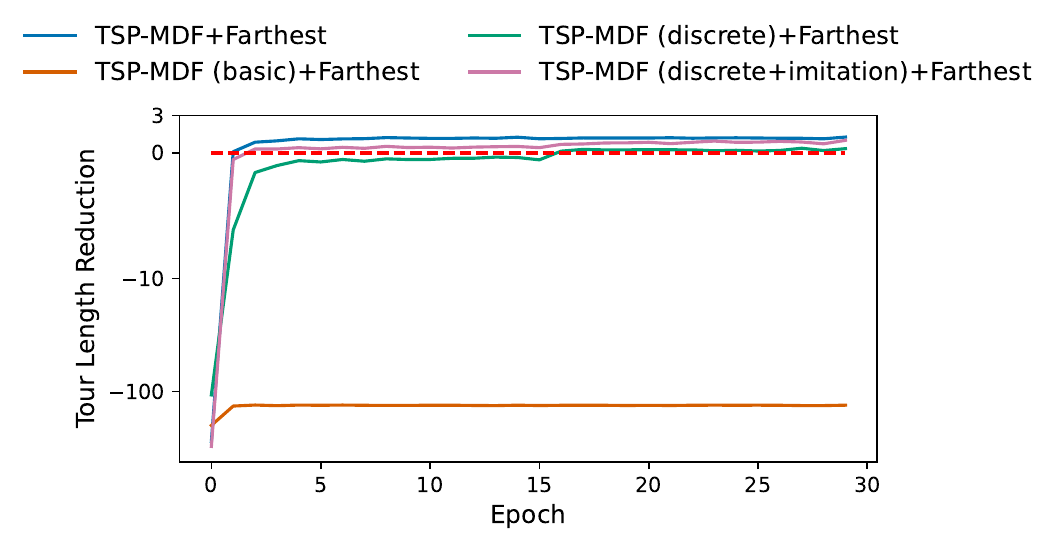}
        \caption{TSP-1000 Farthest Insertion} 
    \end{subfigure}
    \hfill
    \begin{subfigure}{0.32\textwidth}
        \centering
        \includegraphics[width=\textwidth]{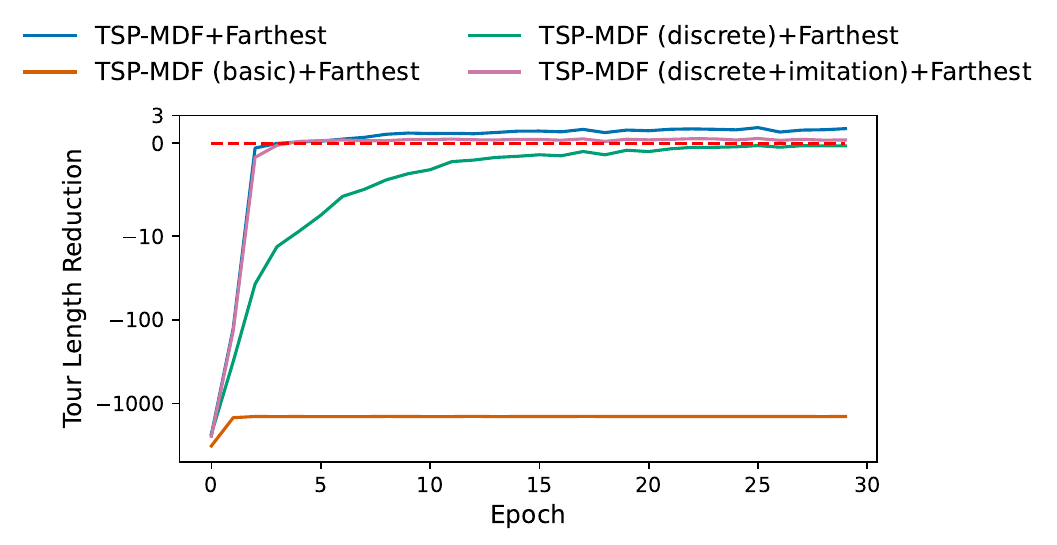}
        \caption{TSP-10000 Farthest Insertion} 
    \end{subfigure}
    \caption{Ablation study: average tour length reduction on training TSP instances.}
    \label{fig:ablation}
    
    \end{minipage}
\end{figure*}

\subsection{Ablation Study}
\begin{table}[t]
\caption{Ablation study on test instances.}
\vspace{-0.3cm}
    \centering
    \resizebox{\linewidth}{!}{\begin{tabular}{c|c|c|c}
         \toprule
         Method&TSP-500&TSP-1000&TSP-10000\\
         \midrule
         Nearest Insertion&$20.62$&$28.95$&$90.49$\\
         TSP-MDF (random) (T=30) + Nearest&$20.62/203.69$&$28.95/416.06$&$90.49/4325.00$\\
         TSP-MDF (basic) (T=30) + Nearest&$20.62/86.15$&$28.95/162.85$&$90.49/1518.46$\\
         TSP-MDF (discrete) (T=30) + Nearest&$20.55/20.75$&$28.92/29.24$&$90.13/90.13$\\
         TSP-MDF (discrete + imitation) (T=30) + Nearest&$19.21/19.21$&$27.53/27.53$&$89.41/89.41$\\
         TSP-MDF (T=30) + Nearest&$18.94/18.94$&$27.09/27.09$&$88.86/88.86$\\

         \midrule
         Farthest Insertion&$18.29$&$25.73$&$80.69$\\
         TSP-MDF  (random) (T=30) + Farthest&$18.29/203.26$&$25.73/415.60$&$80.69/4323.17$\\
         TSP-MDF (basic) (T=30) + Farthest&$18.29/82.30$&$25.73/157.53$&$80.69/1505.33$\\
         TSP-MDF (discrete) (T=30) + Farthest&$18.01/18.03$&$25.52/25.52$&$80.67/80.81$\\
         TSP-MDF (discrete + imitation) (T=30) + Farthest&$17.55/17.55$&$24.68/24.68$&$80.26/80.26$\\
         TSP-MDF (T=30) + Farthest&$17.26/17.26$&$24.44/24.44$&$78.91/78.91$\\
         \bottomrule
    \end{tabular}}
    \label{tab:ablation}
    
\end{table}

\noindent\textbf{Basic design of neural-based instance modifier.}\quad To evaluate the performance improvements brought by the key designs of the neural-based instance modifier, we start from a base model, TSP-MDF (basic), which adopts only the basic design described in Section~\ref{sec:basic}. Table~\ref{tab:ablation} shows the performance of neural-based instance modifier on test instances, and Figure~\ref{fig:ablation} presents the tour length reduction on the training instances. It is clear to see that although this basic neural-based instance modifier outperforms random modification (i.e., TSP-MDF (random)), it still struggles to identify meaningful modifications, thus fails to help the base tour constructor to find better tours on both training and test instances.

\vspace{1mm}\noindent\textbf{Discretization of coordinate offset.}\quad To evaluate the effectiveness of coordinate offset discretization, we introduce a variant model, TSP-MDF (discrete), which extends the basic neural-based instance modifier by discretizing the coordinate offsets as described in Section~\ref{sec:discrete}. As shown in Table~\ref{tab:ablation} and Figure~\ref{fig:ablation}, the proposed discretization significantly improves performance compared to the basic design with continuous offsets. In particular, the neural-based instance modifier begins to identify meaningful modifications that effectively guide the base tour constructor toward better tours on both training and test instances. This improvement can be attributed to the fact that discretizing coordinate offsets reduces the search space to a finite set, while the resulting Categorical distribution enables more flexible yet stable sampling. However, discretizing the coordinate offsets alone is still insufficient for the neural-based instance modifier to identify high-quality modifications that can substantially enhance the base tour constructor.

\vspace{1mm}\noindent\textbf{Self-imitation learning.}\quad To further improve the neural-based instance modifier, we introduce self-imitation learning into its training. To evaluate this design, we create a variant model, TSP-MDF (discrete + imitation), which extends TSP-MDF (discrete) by additionally applying self-imitation learning as described in Section~\ref{sec:imitation}. As shown in Figure~\ref{fig:ablation}, self-imitation learning enables the neural-based instance modifier to quickly discover meaningful modifications that yield positive tour length reductions on training instances, even at the early stage of training within only a few epochs. Thus, the neural-based instance modifier can sample high-quality modifications that significantly enhance the performance of the base tour constructors on test instances, as reported in Table~\ref{tab:ablation}.

This improvement arises because, when relying solely on REINFORCE (i.e., Eq.~\ref{eq:reinforce}), the neural-based instance modifier must explore the search space randomly at cold start and update its parameters based only on these randomly generated modified instances. However, as demonstrated in Section~\ref{sec:ablation_rand} (i.e., TSP-MDF (random)), such random modifications can severely distort the original instances and result in tours of very poor quality. As a result, training the instance modifier purely on these random instances is highly inefficient during the cold-start phase, leading to substantial training time being wasted on unpromising regions of the search space.

By leveraging the self-imitation learning strategy, we anchor the exploration process to relatively promising regions of the search space, thereby preventing the instance modifier from sampling destructive random modifications. This substantially improves training efficiency and enables the neural-based instance modifier to rapidly learn how to generate high-quality instance modifications.

\vspace{1mm}\noindent\textbf{Unification of target output space.}\quad
To evaluate the necessity of unifying the target output space across nodes, we compare TSP-MDF with TSP-MDF (discrete + imitation). The latter can be viewed as a variant of TSP-MDF that directly predicts the distribution of refined coordinate offsets for each node, i.e., without unifying the target output space across nodes. As shown in Figure~\ref{fig:ablation} and Table~\ref{tab:ablation}, TSP-MDF achieves higher training efficiency and yields greater tour length reductions on test instances compared to TSP-MDF (discrete + imitation).
This performance gap arises because, when directly predicting refined coordinate offsets, the instance modifier must not only learn how to refine the previous offsets but also implicitly infer the magnitude of those previous offsets. As a result, the target output space varies substantially across nodes, increasing the difficulty of training for both the reinforcement learning and self-imitation learning. In contrast, by unifying the target output space and predicting only the distribution of coordinate offset refinement, TSP-MDF simplifies and stabilizes the training. Therefore, unifying the target output space is essential for efficient and effective training of the neural-based instance modifier.

\begin{table*}[ht]
\caption{Results on TSPLib. Each instance modification iteration of TSP-MDF samples $100$ modified instances. ``OOM'' indicates that the method ran out of GPU memory ($24$ GB). }
    \centering
    \resizebox{0.8\linewidth}{!}{\begin{tabular}{l|cc|cc|cc}
        \toprule
        \multirow{2}*{Method}
        &\multicolumn{2}{c|}{TSPLIB($1-500$)}&\multicolumn{2}{c|}{TSPLib($501-1000$)}&\multicolumn{2}{c}{TSPLib($1001-10000$)}\\
        &Length $\downarrow$&Inference Time $\downarrow$&Length $\downarrow$&Inference Time $\downarrow$&Length $\downarrow$&Inference Time $\downarrow$\\
        \midrule
        DIMES&$34422.59$&$3.16$s&$2250053.44$&$2.41$s&$1336129.23$&$24.72$s\\
        DIMES+AS&$31314.18$&$5$m$44$s&$2069587.26$&$4$m$8$s&$1260123.03$&$21$m$41$s\\
        DeepACO&$33611.65$&$2$m$24$s&$2391352.04$&$1$m$40$s&OOM&OOM\\
        Pointerformer&$31370.40$&$13.30$s&$2265651.43$&$12.58$s&OOM&OOM\\
        \midrule
        DIFUSCO&$31995.12$&$1$m$48$s&$2155408.74$&$2$m$16$s&$1682753.14$&$22$m$31$s\\
        FastT2T&$32757.16$&$8.36$s&$2090940.55$&$5.43$s&$1577897.58$&$2$m$9$s\\
        FastT2T+MS+GS&$30953.51$&$1$m$4$s&$2025285.39$&$34.43$s&OOM&OOM\\
        \midrule
        Nearest Insertion&$35896.11$&$0.01$s&$2321590.17$&$0.017$s&$1365494.55$&$0.67$s\\
        TSP-MDF ($T=1$) + Nearest&$34833.57$&$0.32$s&$2312856.72$&$0.21$s&$1349780.89$&$8.47$s \\
        TSP-MDF ($T=30$) + Nearest&$33040.38$&$10.68$s&$2198357.65$&$9.10$s&$1330555.17$&$4$m$27$s \\
        \hdashline
        Farthest Insertion&$32051.45$&$0.01$s&$2089399.54$&$0.014$s&$1274799.74$&$0.64$s\\
        TSP-MDF ($T=1$) + Farthest&$31440.27$&$0.31$s&$2078129.38$&$0.19$s&$1264057.85$&$8.96$s \\
        TSP-MDF ($T=30$) + Farthest&$\textbf{30468.03}$&$10.48$s&$\textbf{1995797.52}$&$8.73$s&$\textbf{1238924.88}$&$4$m$26$s \\
        \bottomrule
        
    \end{tabular}}
    \label{tab:tsplib}
\end{table*}

\subsection{Generalization Ability to Real-world Benchmarks}
\label{appendix:tsplib}
To evaluate the generalization ability of TSP-MDF on real-world datasets, we consider the neural-based instance modifier trained solely on random instances (i.e., the training sets of TSP-$n$ datasets) and test on the well-known real-world benchmarks TSPLib~\cite{reinelt1991tsplib} with various distributions. We use all $87$ 2-D TSPLib instances with up to $10000$ nodes, and group them according to instance size into three subsets: TSPLIB($1-500$), TSPLib($501-1000$), and TSPLib($1001-10000$).
Specifically, we evaluate TSP-MDF trained on random instances of $500$, $1000$ and $10000$ nodes for TSPLIB($1-500$), TSPLib($501-1000$), and TSPLib($1001-10000$), respectively.

Table~\ref{tab:tsplib} presents the performance results on TSPLib. The results clearly demonstrate that TSP-MDF consistently enhances traditional heuristic tour constructors, even when evaluated on test instances whose distributions differ from those of the training instances. In particular, while Farthest Insertion originally performs worse than some neural-based heuristic tour constructors, the TSP-MDF-enhanced Farthest Insertion consistently outperforms all neural-based heuristic tour constructors across all three TSPLib subsets, while maintaining high computational efficiency. Notably, DIMES+AS is fine-tuned on each individual test instance via Active Search~\cite{hottung2021efficient}, yet it still fails to outperform our TSP-MDF-enhanced Farthest Insertion. Overall, these results demonstrate the strong generalization capability of TSP-MDF on real-world TSP instances.

\subsection{Visualization}
\label{appendix:visualization}
\begin{figure*}[htbp]
\centering
    \resizebox{0.8\linewidth}{!}{\begin{minipage}{\textwidth}
        
    % 第一行
    \begin{subfigure}{0.32\textwidth}
        \centering
        \includegraphics[width=\textwidth]{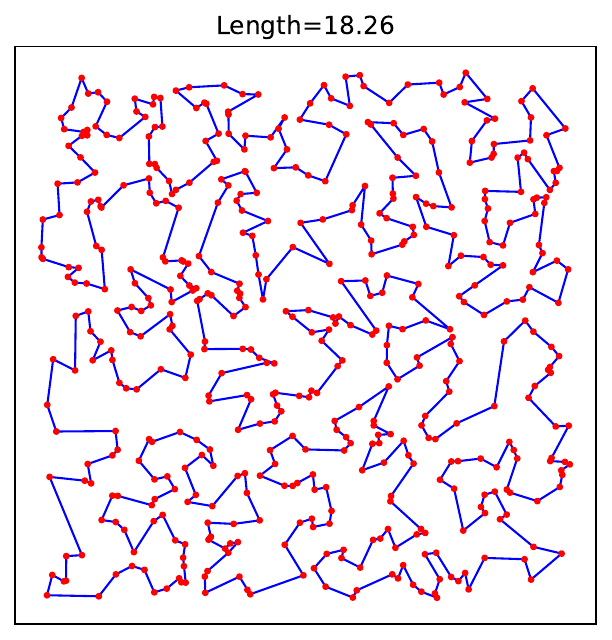}
        \caption{Tour approximated by the Farthest Insertion on the original instance.}
    \end{subfigure}
    \hfill
    \begin{subfigure}{0.32\textwidth}
        \centering
        \includegraphics[width=\textwidth]{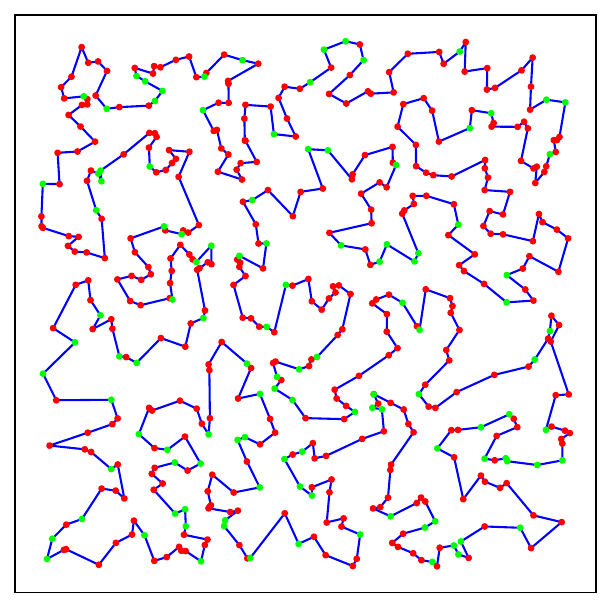}
        \caption{Tour approximated by the Farthest Insertion on the modified instance.} 
    \end{subfigure}
    \hfill
    \begin{subfigure}{0.32\textwidth}
        \centering
        \includegraphics[width=\textwidth]{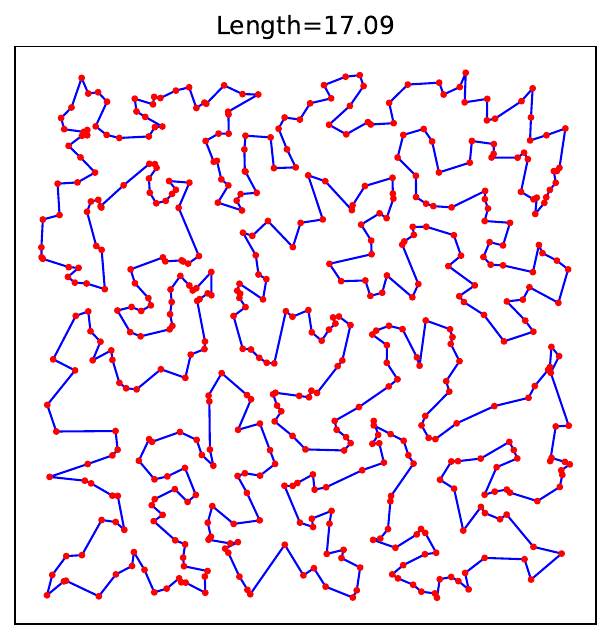}
        \caption{Tour approximated in (b), mapped back to the original instance.} 
    \end{subfigure}

    \vskip\baselineskip
      \begin{subfigure}{0.32\textwidth}
        \centering
        \includegraphics[width=\textwidth]{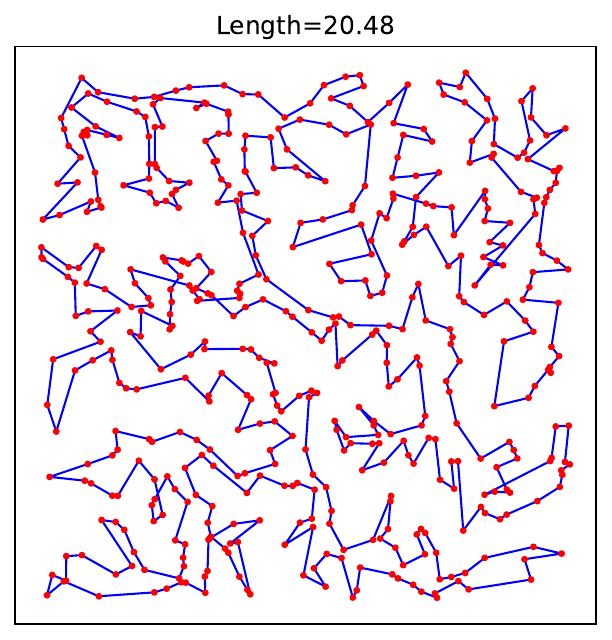}
        \caption{Tour approximated by the Nearest Insertion on the original instance.}
    \end{subfigure}
    \hfill
    \begin{subfigure}{0.32\textwidth}
        \centering
        \includegraphics[width=\textwidth]{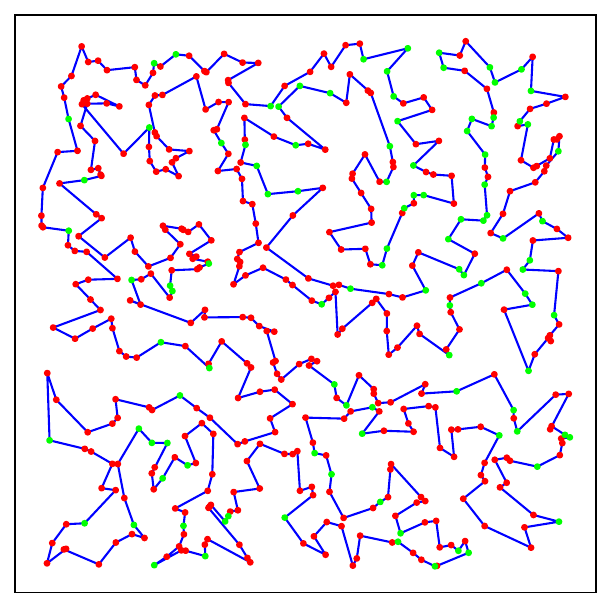}
        \caption{Tour approximated by the Nearest Insertion on the modified instance.} 
    \end{subfigure}
    \hfill
    \begin{subfigure}{0.32\textwidth}
        \centering
        \includegraphics[width=\textwidth]{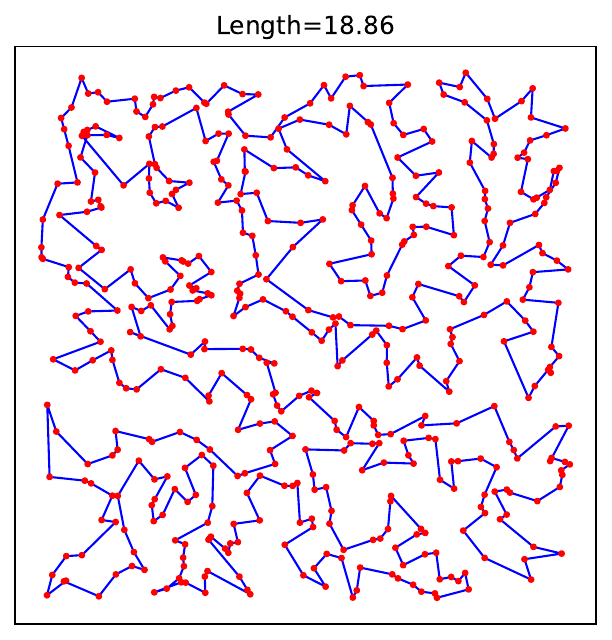}
        \caption{Tour approximated in (e), mapped back to the original instance.} 
    \end{subfigure}
    
    \end{minipage}}
     \caption{An example of TSP-MDF-enhanced heuristic tour constructors on a TSP-500 test instance. Green nodes in the modified instance indicate nodes whose coordinates have been modified.}
    \label{fig:visualization}
\end{figure*}

We visualize how traditional heuristic tour constructors are enhanced by our TSP-MDF framework. Figure~\ref{fig:visualization} presents a representative example on a TSP-500 test instance. Specifically, Figures~\ref{fig:visualization}(a) and (d) show the tours produced by the base heuristic tour constructors on the original instances, while Figures~\ref{fig:visualization}(b) and (e) illustrate the tours obtained on the modified instances. Figures~\ref{fig:visualization}(c) and (f) further visualize the tours from the modified instances after mapping them back to the original instances. By slightly perturbing the node coordinates, TSP-MDF enables traditional heuristic tour constructors to explore substantially different solution trajectories, helping them escape poor local optima and ultimately approximate significantly shorter tours.

\end{document}